\documentclass[runningheads]{llncs}
\usepackage{graphicx}
\usepackage{comment}
\usepackage{amsmath,amssymb,amsfonts}
\usepackage{adjustbox}
\usepackage{booktabs}
\usepackage{hyperref}

\usepackage[linesnumbered,ruled,vlined]{algorithm2e}

\SetKwInput{KwInput}{Input}                
\SetKwInput{KwOutput}{Output}              

\begin{document}
\title{TS-CausalNN: Learning Temporal Causal Relations from Non-linear Non-stationary Time Series Data}

\titlerunning{TS-CausalNN}

\author{Omar Faruque \inst{1} \orcidID{0009-0006-8650-4366} \and
Sahara Ali\inst{1} \and Xue Zheng\inst{2} \and Jianwu Wang \inst{1}\orcidID{0000-0002-9933-1170}}
\authorrunning{Faruque, et al.}
%
%
\institute{Department of Information Systems, University of Maryland, Baltimore County, Baltimore, MD, United States \\
\email{\{omarfaruque, sali9, jianwu\}@umbc.edu}\\
\url{https://bdal.umbc.edu} \and
Climate Science Section, Lawrence Livermore National Laboratory, Livermore, CA, United States\\
\email{zheng7@llnl.gov}}
%
\maketitle              
\begin{abstract}
  The growing availability and importance of time series data across various domains, including environmental science, epidemiology, and economics, has led to an increasing need for time-series causal discovery methods that can identify the intricate relationships in the non-stationary, non-linear, and often noisy real world data. However, the majority of current time series causal discovery methods assume stationarity and linear relations in data, making them infeasible for the task. Further, the recent deep learning-based methods rely on the traditional causal structure learning approaches making them computationally expensive. In this paper, we propose a Time-Series Causal Neural Network (TS-CausalNN) - a deep learning technique to discover contemporaneous and lagged causal relations simultaneously. Our proposed architecture comprises (i) convolutional blocks comprising parallel custom causal layers, (ii) acyclicity constraint, and (iii) optimization techniques using the augmented Lagrangian approach. In addition to the simple parallel design, an advantage of the proposed model is that it naturally handles the non-stationarity and non-linearity of the data. Through experiments on multiple synthetic and real world datasets, we demonstrate the empirical proficiency of our proposed approach as compared to several state-of-the-art methods. The inferred graphs for the real world dataset are in good agreement with the domain understanding. 
  

\end{abstract}
\section{Introduction}
\label{intro}
Time series data generated by different natural systems, such as climate and environment, can possess different features like non-linearity, non-stationarity, presence of different noise categories, and autocorrelation, which are mostly multivariate \cite{Runge2019InferringCF}. These intricate characteristics of time series data ingraft complex challenges in understanding the dependencies of different components of these natural systems. One popular approach to simplify the understanding of large multivariate time series datasets is to graphically represent the data generation model using directed acyclic graphs (DAGs), which is a very convenient way to express complex systems in a highly interpretable manner and also provide causal insights into the underlying processes of the system \cite{pearl2000models}. DAG representation of a system plays a vital role in decision-making and future condition prediction in different applications like causal inference \cite{Pearl1991ProbabilisticRI,spirtes2000causation}, neuroscience \cite{rajapakse2007learning}, medicine \cite{medicine}, economics \cite{economics}, finance \cite{finance}, and machine learning \cite{ml-application}. Learning DAG through causal discovery from observational time series data is very challenging when controlled experiments with different population sub-groups are not possible or unethical \cite{spirtes2000causation,peters2017elements}.  

Several state-of-the-art methods have been developed for causal discovery from temporal data based on constraint-based and score-based methodologies. Constraint-based causal discovery methods \cite{pcmci,pcmci+,lpcmci,ts-fci,cd-nod} learn the conditional independencies from the dataset using different tests and construct the DAG to reflect this independence knowledge. One major limitation of the conditional independence test is that it requires a large number of observation samples to generate reliable test scores \cite{Shah_2020}. Score-based causal discovery methods use a score function to quantify the predicted causal graph based on the adjacency matrix and try to optimize the score function gradually by enforcing the acyclicity constraint of the graph. The large search domain of the score function based on the adjacency matrix combination makes the optimization process very challenging and sometimes requires additional knowledge of the DAG. The combinatorial characteristic of the score optimization transformed into a continuous optimization problem by Zheng et al. \cite{zheng2018dags} formulating an equivalent acyclicity constraint using the trace exponential of the predicted adjacency matrix. Through this continuous formulation of the acyclicity constraint now causal graphs can be optimized using several widely used gradient-based optimizers. This drastically changes the domain of multivariate time series causal discovery using neural network-based approaches and several methods have been proposed in recent years.


Time series causal discovery leveraging the power of neural networks has gained much attention as an active area of research in several fields. However, the majority of existing methodologies are designed under the assumption of stationarity, a condition often violated in real world scenarios where dynamic systems evolve periodically \cite{hasan2023survey,runge2019inferring}. Some existing approaches also require prior knowledge of the domain model like linearity, noise distribution, and parametric information. Motivated by the tremendous success of neural networks \cite{gu2018recent}, in this paper, we propose Time-Series Causal Neural Network (TS-CausalNN) - a novel causal discovery method from temporal data using a convolution neural network. Where the causal relationship between each child to its temporal parents is learned by using a custom 2D convolution layer. Our proposed model is capable of capturing the causal structure from multivariate temporal data without any noise and data distribution assumptions. Our method is designed to be applicable across diverse domains that scale gracefully for multivariate time series data. The contributions of this paper are three-fold. First, we propose a 2D convolutional neural network layer to learn the causal relationships from the multivariate time series dataset. By utilizing the power of the CNN, our proposed model can handle both the stationary and non-stationary data for linear and non-linear structural models with the presence of different noise distributions. The proposed parallel network architecture has not been used earlier. Second, the simplified optimization routine of the proposed model can identify lagged and contemporaneous causal links simultaneously. The integration of the acyclicity constraint and sparsity penalty into the optimization process helps to learn better causal graphs. Finally, we conduct extensive evaluations of the proposed model with state-of-the-art methods using synthetic and real world datasets. The proposed model achieves better evaluation scores for generated causal discovery graphs for most of the cases, making TS-CausalNN a strong contender for time-series causal discovery.

The remainder of this paper is organized as follows. Section 2 describes the preliminaries of temporal casual discovery and provides an overview of existing literature highlighting key methods and their limitations. In section 3 we describe the proposed method in detail, emphasizing its key components and innovations. Section 4 presents the datasets, preprocessing steps, and experimental configurations used to evaluate the proposed method. The quantitative and qualitative comparison of the performance of our method with baseline approaches and real world applications are discussed in section 5.
Section 6 summarizes the main contributions of the paper, acknowledges limitations, and suggests directions for future work.

\section{Preliminaries}
\begin{figure}[h]
  \centering
  \includegraphics[width=\linewidth]{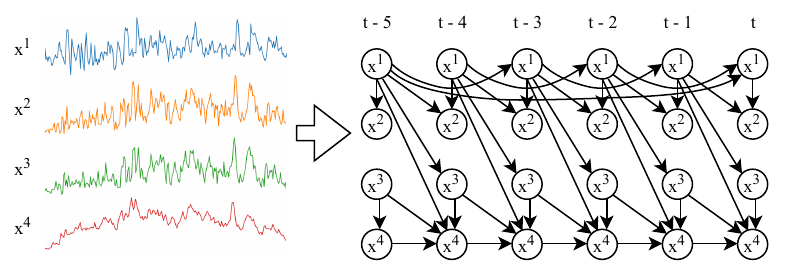}
  \caption{Temporal causal graph learned (right) from multivariate time series data (left) with causal links from the same and previous timesteps. Each node in the graph represents one variable at a specific timestep ( $t$ is present time and $t-i$ is a previous timestep). A directed edge denotes a causal relationship between cause and effect.}
\label{full-causal-graph}
\end{figure}

Temporal causal discovery learns directed acyclic graphs (DAG) from time series data. Each directed edge of the DAG represents the influence of the cause variable on the target variable. As the order of data is strictly maintained in the time series case, influences from cause variables to the target variable can only come from the same and previous timesteps, also called temporal precedence (the right side of Figure~\ref{full-causal-graph}). This feature makes causal graph learning from time series data more challenging compared to independent and identically distributed (IID) data.

Let's consider a multivariate time series dataset $X = \{x^1, x^2, x^3, …, x^n\}$ consisting of $n$ variables, and each variable is measured for $T$ timesteps. Each variable $x^i$ at a specific time point $t$ can be caused by other variables in the same time point ($t$) and all variables from previous timesteps ($0$ to $t-1$). The effect from previous timesteps, also called lagged effects, can propagate from infinite earlier time points, but for DAG learning purposes we will consider a maximum time lag, $l_{max}$. So the set of possible cause variables of each time series $x^i$ at time $t$ is $PA_{x^i} \in [ \{X_{(t-l_{max})}, X_{(t-l_{max}-1)}, ..., X_{(t-1)}, X_t\} - x^i]$. The goal is to learn a causal graph $G(V, E)$ from the provided dataset such that its vertices resemble time-lagged and current time variables in the time series and its directed edges express causal links from parents to target variables. So the vertices and edges are denoted as \sloppy{ $V = \{X_{(t-l_{max})}, X_{(t-l_{max}-1)}, ..., X_{(t-1)}, X_t\}$, $E = \{(V_i,V_j): V_i,V_j \in \{X_{(t-l_{max})}, X_{(t-l_{max}-1)}, ..., X_{(t-1)}, X_t\}\}$, respectively.} Let the weighted adjacency matrix of causal graph G be denoted by $W \in R^{(n \times (l_{max}+1)) \times n}$, which contains both the time-lagged and instantaneous part of the causal links. The structural equation model of the time series can be defined as:
\begin{equation}
\begin{split}
       X_t = f_W(X_{(t-l_{max})}, X_{(t-l_{max}-1)}, ..., X_{(t-1)}, X_t) + e_t 
\end{split}
   \label{eq:sem}
\end{equation}

\noindent , where the noise term $e_t$ can be of any type, independent of cause and effect, and the structural function $f_W()$ can be any linear or non-linear data generation process.


In the time-lagged part of the adjacency matrix $W$, the nodes from previous $l_{max}$ timesteps of $t$ will always be the source/cause node of the lagged causal links. So, we do not need to explicitly focus on the acyclicity of the time-lagged part of $W$. However, for the contemporary part of the $W$ at $t$, each node can serve as both the source and the target of the causal links where we have to maintain the acyclicity of the DAG. 

Several methods can be employed to recover the adjacency matrix $W$ of the causal graph from time series data, like search-based, constraint-based, score-based, graph-based methods, etc. To formulate this combinatorial search problem into a continuous optimization nature Zheng et al. \cite{zheng2018dags} proposed an algebraic representation of the acyclicity constraint of DAG with the adjacency matrix exponential. This new representation of acyclicity opens a window to learning the DAG using generic continuous optimization routines like neural networks by minimizing the score. However, simultaneously learning the lagged and contemporaneous part of the adjacency matrix is very challenging for complex datasets. Since any variable might be the cause of another effect variable, cycles can occur in the contemporaneous part of the adjacency matrix. Consequently, these two parts of the adjacency matrix require a different set of optimization criteria.

\section{Proposed Methodology}
\label{proposed_method}
We can consider the causal graph generating task as an unsupervised learning process of the adjacency matrix $W$ given the multivariate time series data $X = \{x^1, x^2, x^3, …, x^n\}$ of $T$ observations. To learn the directed adjacency matrix $W$ of the temporal causal graph $G$, we propose a neural network-based unsupervised model. The proposed time series causal neural network model will learn the instantaneous $(X_t \to X_t)$ and time-lagged $(\{X_{(t-l_{max})}, X_{(t-l_{max}-1)},..., X_{(t-1)}\} \to X_t)$ causal links of $W$ for maximum time lag ($l_{max} > 0$) in the same gradient propagation using a single neural network. To perform the temporal causal learning, we propose a causal convolution 2D layer.

\subsection{Time Series Causal Neural Network}
\label{pro-model}
\begin{figure}[h!]
  \centering
  \includegraphics[width=\linewidth]{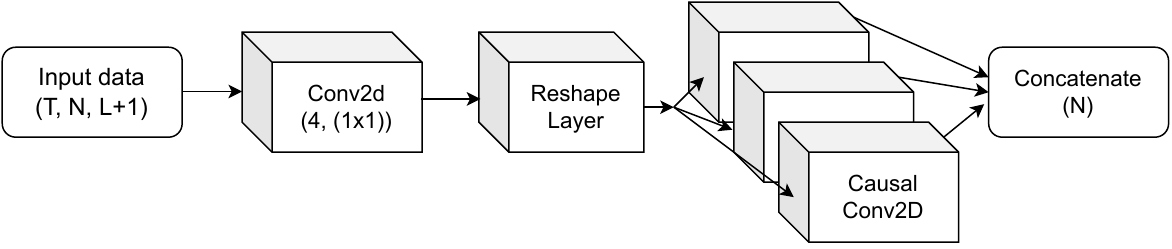}
  \caption{Proposed TS-CausalNN model architecture to learn full temporal causal graph. The graph can be learned from the parallel Causal Conv2d layers.}
  \label{model-arch}
\end{figure}

As illustrated in Figure~\ref{model-arch}, our proposed TS-CausalNN model consists of two separate blocks of 2-dimensional convolution layers, which helps its simplicity and efficiency. The first 2D convolution layer of the model transforms input data into a latent representation, which helps the model to learn any non-linear relationship present in the data generation process. The Causal Conv2D block contains $n$ number of parallel custom causal layers to learn the time-lagged and instantaneous causal relationships of each input variable to its parent variables. Following a similar analogy used by Zheng et al. \cite{notears-mlp}, DAG-GNN \cite{yu2019dag} and NOTEARS-MLP \cite{zheng2018dags}, we will learn the causal links from the parameters of the Conv2D block. Here the links learned by this causal block are always unidirectional and the parallel blocks help to learn the causal links for each input variable independently. Finally, the results of each parallel causal layer are aggregated to generate the model output, which will be used in the optimization process of the learned causal graph.  


\begin{figure}[h!]
  \centering
  \includegraphics[width=\linewidth]{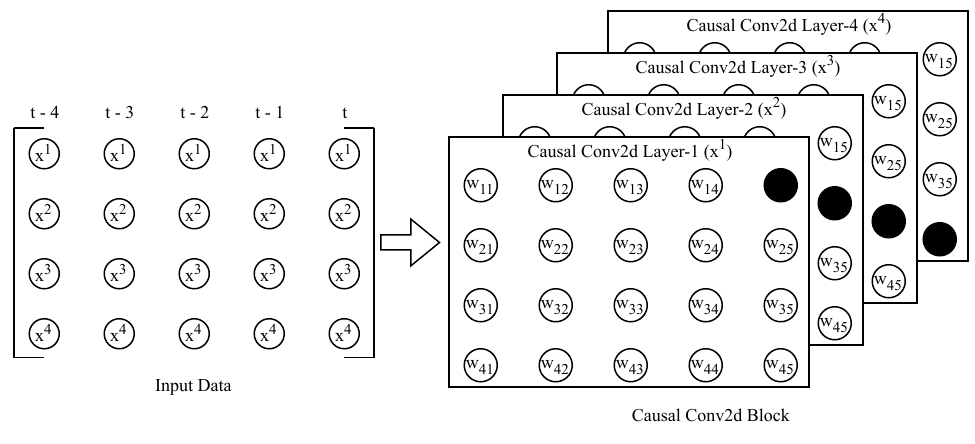}
  \caption{Proposed custom Causal Conv2D Layers.}
  \label{causal-conv-layer}
\end{figure}

Our proposed Causal Convolution 2D (Causal Conv2D) layer takes input in a similar structure as shown in the full causal graph on the right side of Figure \ref{full-causal-graph}, with lagged data followed by the current time point’s data.  Each Causal Conv2D layer is designed to learn the causal links of an input variable for example $x^1$ from all possible parents \sloppy{ $PA_{x^1} \in \{ x^1_{(t-l_{max})},  x^1_{(t-l_{max}-1)},...,  x^1_{(t-1)}, x^2_{(t-l_{max})}, x^2_{(t-l_{max}-1)},..., x^2_{(t-1)}, x^2_t,$ $ ..., x^n_{(t-l_{max})}, x^n_{(t-l_{max}-1)},..., x^n_{(t-1)}, x^n_t\}$ of that variable.} The variable itself cannot be included in the set of its parent variables. Let's assume we have a time series dataset with 4 variables $X = \{x^1, x^2, x^3, x^4\}$, and for lagged effects consider the maximum time lag $l_{max} = 4$. So the input data will be a matrix of size $(4 \times 5)$ one row for each variable and $l_{max} + 1 = 5$ column for lagged and contemporaneous variables. To learn the temporal causal graph for these 4 variables, as shown in Figure~\ref{causal-conv-layer}, we have to employ 4 parallel Causal Conv2D layers one layer for each variable. Each of these layers predicts the expectation of the target variable at the current timestep $t$ given all lagged and instantaneous parents (Equation \ref{eq:prediction}). To exclude the target variable from the parents list, its weight is set to zero.      


\begin{equation}
   E[x^i|PA_{x^i}] = f_{W^{x^i}}(PA_{x^i})
   \label{eq:prediction}
\end{equation}

Here $f_{W^{x^i}}()$ is the function learned at the Causal Conv2D layer considering $x^i$ as the target variable and $W^{x^i}$ is the set of weight parameters of that layer. Motivated by NOTEARS-MLP \cite{zheng2018dags} and DAG-GNN \cite{yu2019dag}, we derive the adjacency matrix of causal DAG from the weight parameters of the Causal Conv2d layers. The value of the weight parameter of the layer for a target variable represents the strength of the causal link from its parent. The weight parameter $W^{x^k}_{ij} = 0$ means the target variable $x^k$ is independent of the cause variable $x^i$ at timestep $j$ in the $PA_{x^1}$ list. If the $W^{x^k}_{ij} > 0$ then the variable $x^k$ has a causal edge from the parent variable $x^i$ at a specific time lag $j$. After learning the final weight parameters of the target variable, we apply a thresholding operation to prune causal links with weak dependency strength, $W^{x^k}_{\omega} = (W^{x^k} > \omega)$, where $\omega$ is the threshold value. Finally, the weight parameters of all variables after thresholding are concatenated to generate the combined adjacency matrix of the final causal graph.

\subsection{Acyclicity Constraint}
Maintaining the acyclic property of the adjacency matrix makes it very difficult to learn causal graphs from time series data. Using a continuous optimization process like a neural network to learn the causal graph does not guarantee the acyclicity of the learned adjacency matrix of the DAG. To impose the acyclicity restriction in the adjacency matrix of the learned causal DAG, we will use a similar equality constraint $h(W) = 0$ as Zheng et al. \cite{zheng2018dags}. The function $h(W)$ is defined using the trace exponential of the elementwise product of the adjacency matrix with itself. 

\begin{equation}
   h(W) = tr \: e^{W o W} - n
   \label{eq:acyclicity-m}
\end{equation}
Here $n$ is the number of variables in the dataset. We cannot use the learned adjacency matrix $W$ of the proposed TS-CausalNN method directly in this equality function because $W$ contains both the time-lagged $(t-l_{max}, t-l_{max}-1, …,  t-1)$ and contemporaneous $(t)$ edges of the causal graph. The time-lagged edges of the causal graph will always go forward in time from the previous timesteps to the current timestep $t$, therefore we do not need to compute acyclicity for this part of the graph. So we have to apply the acyclicity constraint on the contemporaneous part of the adjacency matrix, $W^t$. Hence the function can be stated below and the equality will be satisfied if and only if the contemporaneous part of the adjacency matrix ($W^t$) is acyclic.
\begin{equation}
   h(W^t) = tr \: e^{W^t o W^t} - n = 0
   \label{eq:acyclicity}
\end{equation}

\subsection{Optimization}
The goal of the proposed causal discovery method is to learn the adjacency matrix $W$ of the DAG from the given time-series dataset $X$. The adjacency matrix $W$ contains the edges from the time-lagged variables together with the contemporaneous variables to the target variables. So the proposed  TS-CausalNN model has to estimate the target variables using all possible parents and find the causal influence ($W$) of each parent on the target child, where the contemporaneous part of $W^t$ must satisfy the acyclicity requirement. So the optimization objective of the model is to minimize the least-square loss ($L$) of the target variables with the acyclicity constraint on $W^t$.  
\begin{equation}
    \min_{W} L(W) \; \; \mathrm{subject \; to} \; \; h(W^t) = 0 \; \; \mathrm{for \; acyclicity}
    \label{eq:objective-fn}
\end{equation}
\begin{equation}
\mathrm{where} \; \;  L(W) = \frac{1}{T}|| X - WX||^2_F 
    \label{eq:ls-loss}
\end{equation}
\begin{equation}
       h(W^t) =  tr(e^{W^t o W^t}) - n = 0
    \label{eq:acyclicity-cons}
\end{equation}

The function $ h(W^t)$ will be equal to $0$ if and only if the corresponding sub-graph graph of matrix $W^t$ does not have any cycle. We cannot directly integrate the equality-constraint problem of the acyclicity into the continuous optimization problem. But the equality constraint $h(W^t) = 0$ can be solved using continuous optimization after converting this into an unconstraint subproblem \cite{zheng2018dags}. Therefore we will use the augmented Lagrangian method to solve the equality constraint problem. An additional penalty is incorporated with the objective function to enforce the sparsity of the adjacency matrix using the $L1$ norm of $W$.

The least-square loss and the acyclicity constraint (Equations \ref{eq:ls-loss} and \ref{eq:acyclicity-cons}) will try to increase the weight parameter to minimize the loss. On the contrary, the L1 norm of W will try to reduce weight values to zero to keep a minimum number of non-zero entries. By working contrariwise to each other the loss function will be optimized in an equilibrium way. Hence the final unconstrained objective function is:
\begin{equation}
       \min_{W} \left [ L(W) + \frac{\rho}{2}|h(W^t)|^2 + \alpha h(W^t) + \lambda ||W||_1 \right ]
    \label{eq:objective-total}
\end{equation}

\noindent , where 2nd and 3rd terms are the augmented Lagrangian for the acyclicity constraint, $\alpha$ is the Lagrange multiplier, $\rho > 0$ is the penalty parameter of the augmented Lagrangian, and $\lambda$ is the sparsity penalty parameter. Here $\lambda$ is a model hyperparameter and we set it at $0.1$ for this study. This objective function can be minimized using any state-of-the-art continuous optimizer. An excellent characteristic of the augmented Lagrangian approach is its ability to accurately approximate the solution of a constrained problem using the solution of unconstrained problems by gradually increasing the penalty parameter $\rho$ but not to infinity. Hence we have to gradually increase $\rho$ to minimize the unconstrained augmented Lagrangian and the value of the $\alpha$ also be updated accordingly. The rule for updating $\rho$ and $\alpha$ based on the value of $h(W^i)$ is given by:

\begin{equation}
       \rho^{i+1} = \left\{\begin{matrix}
(1+ \beta) \rho^{i}, & \textup{if} \; \; h(W^i) > \gamma h(W^{i-1}) \\ 
\rho^{i}, & \textup{otherwise}
\end{matrix}\right.
    \label{eq:lagrange-1}
\end{equation}
\begin{equation}
       \alpha^{i+1} =  \alpha^{i} +  \rho^{i} h(W^i) 
    \label{eq:Lagrange-2}
\end{equation}
      
\noindent , where $\beta > 0$ and $\gamma < 1$ are hyperparameters, and we find that ($\beta =0.1, \gamma = 0.25$) work better.
The overall process of learning causal DAG is outlined in Algorithm \ref{algorithm1}. 

\begin{algorithm}[!ht]
\DontPrintSemicolon
  \KwInput{Multivariate Time Series Data $X=\{x^1,x^2,x^3,...,x^n\}$ }
  \KwOutput{Adjacency Matrix of the causal graph $W$}
    Instantiate and compile the Causal Neural Network model 
    
    \For{each iteration}
    {
        Train the Neural Network model\;
        Compare the acyclicity loss with the previous iteration\;
        Update the penalty coefficient of the Lagrangian method ($\rho, \alpha$)\;
    }
    Return the adjacency matrix $W$
\caption{TS-CausalNN Algorithm}
\label{algorithm1}
\end{algorithm}

    

\section{Experimental Setup}
\label{experiment}
We mention the dataset and evaluation criterion used for performance comparison in this section.  
Our model is developed on TensorFlow Keras and all experiments are conducted on Google Colab Runtime with CPU for easy reproducibility. Fixed seed value is used for randomized data to make the experimental results reproducible.  

\subsection{Synthetic Datasets}
\begin{figure}[h]
  \centering
  \includegraphics[width=\linewidth]{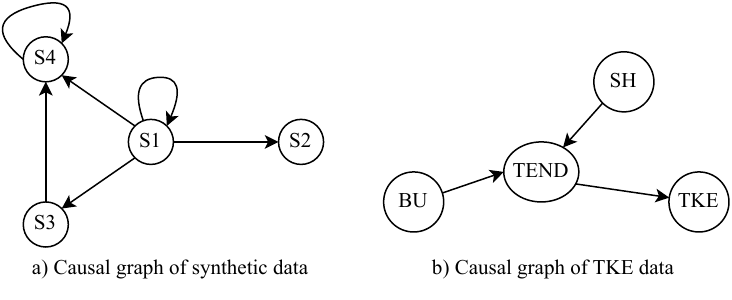}
  \caption{Causal graph of (a) our synthetic datasets and (b) the real world Turbulence Kinetic Energy
(TKE) dataset.}
  \label{true-graph}
\end{figure}

To evaluate the performance of our proposed causal discovery method we have used synthetic datasets. As we know the ground truth causal graph for the synthetic datasets, we can measure and compare the learned causal graph easily. We generate a time series dataset consisting of four non-linear variables using Gaussian white noise $\varepsilon$ (Dataset-1). The mathematical description of each variable is given in Equations 10 to 13. The non-linear characteristic is incorporated in the generation of the synthetic dataset to mimic the dynamic properties of real world natural system data. The corresponding true causal graph for the time-series data is given in Figure \ref{true-graph}a.
    \begin{equation}
        S1_t = 2 \{cos(\frac{t}{10}) + log(|S1_{t-2} - S1_{t-5}| + 1)\} + 0.1 \varepsilon 1 
    \end{equation}
    \begin{equation}
    S2_t = 12 e^{\frac{S1^2_{t-1}}{2}} - 4 e^{\frac{S1^2_{t}}{2}} + \varepsilon 2
    \end{equation}
    \begin{equation}
    S3_t = -10.5 e^{\frac{-S1^2_{t-1}}{2}} + \varepsilon 3
    \end{equation}
    \begin{equation}
    S4_t = -11.5 e^{\frac{-S1^2_{t-1}}{2}} + 13.5e^{\frac{-S3^2_{t-1}}{2}} + 1.2 e^{\frac{-S4^2_{t-1}}{2}} - 5e^{\frac{-S3^2_{t}}{2}} + \varepsilon4
    \end{equation}


We have also generated another version of the synthetic dataset (Dataset-2) from the previous causal graph without the cosine part of the data generation formulas. For this case, we used the Poisson distribution to incorporate random noise into the dataset. The goal is to find the performance and robustness of the causal discovery methods for the time series dataset without having wave-shape-like features with non-Gaussian noise. The formula used to generate this dataset is available in Appendix \ref{syn-data-2}.  

Before applying the proposed causal discovery method, we performed some data preprocessing to ensure the quality and consistency of the input data. Different variables of the dataset have different scales of values. All time series data were normalized within the scale of 0 to 1 to mitigate the impact of scale differences. Then $l_{max}$ previous timesteps data of each current timestep is concatenated in front of it to make the dataset size $(T, n, (l_{max}+1))$. The preprocessed data can be applied to the input convolution 2D layer of the model directly.    

\subsection{Real World Datasets}
\label{real-world-data}
Two real world Earth/atmospheric science datasets, namely Turbulence Kinetic Energy (TKE) and Arctic Sea Ice, were used to evaluate our work. These natural datasets exhibit high variability, non-stationarity, and complex data interactions. Owing to these characteristics, we evaluate our model performance on these datasets to assess how well the proposed model performs on such complex data.

TKE refers to the mean kinetic energy per unit mass associated with eddies in turbulent flow \cite{tke}. It can be measured by the root-mean-square (RMS) velocity fluctuations (unit: $m^2 s^{-2}$). The temporal TKE data used in this study represents the TKE evolution during a typical cumulus-topped boundary layer day (local time 05:00 – 18:00) over the DOE Atmospheric Radiation Measurement (ARM) Southern Great Plains Central Facility. This data file is generated from an idealized numerical simulation using the Weather Research \& Forecasting Model \cite{wrf_skamarock} with modifications from the Large-Eddy Simulation (LES) Symbiotic Simulation and Observation (LASSO) activity, which is developed through the US Department of Energy’s ARM facility \cite{les-arm,endo2015racoro}. Besides TKE, the model also generated the budget terms determining the temporal change in TKE. The major budget terms include the TKE vertical shear production term ($SH, m^2 s^{-3}$), the TKE buoyancy production term ($BU$), and the TKE turbulent and pressure transport term ($TR$). All these budget terms form the net temporal change term of TKE ($TEND, m^2 s^{-3}$). If $TEND$ is positive (negative), TKE will increase (decrease) in the next timestep. Figure~\ref{true-graph}b illustrates how these terms relate to one another through a directed graph.




Arctic sea ice is one of the important components of the world's climate system that has a great impact on the increase of extreme weather events \cite{cohen2014recent,simmonds2014physical,yao2017increased,luo2018changes,luo2019weakened} and the ice is melting rapidly due to various atmospheric conditions \cite{serreze2015arctic,simmonds2015comparing}. Huang et al. \cite{huang2021benchmarking} conducted a causal discovery analysis to uncover the links between the arctic sea ice and the atmosphere. We use the same  11 atmospheric variables with the sea ice extent employed in \cite{huang2021benchmarking}. 
This time series data contains monthly averages from 1980 to 2018 over the Arctic region of 60N. The causal relationship between the atmospheric variables and the sea ice extent based on the physics and microphysics literature review discussed in \cite{huang2021benchmarking} is represented in Appendix \ref{sie-all}.

\subsection{Evaluation Metrics}
To evaluate the performance of the time series causal discovery methods we will use three standard evaluation metrics: Structural Hamming Distance (SHD), F1 Score and False Discovery Rate (FDR). SHD of the directed graph represents the smallest number of edge corrections required to transform the predicted causal graph into the true causal graph. The edges can be corrected by adding new edges, removing existing edges, and changing the direction of an existing edge in the graph. A lower SHD is expected for better performance of the causal discovery method. F1 Score calculate the harmonic mean of precision and recall, providing a balanced measure of performance. The F1 score ranges from 0 to 1 and a higher value means a better prediction of the true graph. FDR explains the rate of predicted wrong edges from all predicted edges considering the direction of each edge. This is the inverse measure of the precision of the prediction.

\section{Results}
In this section, we present the comparative results of the time series causal discovery between the proposed method and state-of-the-art methods. Several typical temporal causal discovery challenges have been considered like time-lagged and contemporaneous causal relations, different noise distributions, nonlinearity of data, and autocorrelation. 

\subsection{Baselines}
To compare the results of the proposed method with baselines we considered PCMCI+\cite{pcmci+}, DYNOTEARS \cite{dynotears}, NTS-NOTEARS \cite{nts-notears}, PCMCI \cite{pcmci}, NOTEARS-MLP \cite{notears-mlp}, and DAG-GNN \cite{yu2019dag}. The first three methods can directly learn causal graphs for time series data. Though the other three methods were proposed for non-temporal data we used these methods due to their popularity and widespread usage in different domains. We transformed the lagged and instantaneous data into a long sequence so that we could apply the transformed dataset to the non-temporal methods to find the lagged and current time causal relationships. For each method, we tuned hyperparameters to get the best evaluation scores.  

\subsection{Quantitative Results}
\label{Quantitative-Results}
To evaluate the performance of the selected state-of-the-art methods we compared the predicted causal graph in both the summary and full temporal graph setting. The qualitative comparison of the baseline methods is reported in Table \ref{comparison-symmary-graph} and Table \ref{comparison-full-graph}. The best results are marked in bold text and underlined values represent the second best score. To get better prediction results, we have applied each method multiple times on the same dataset and reported the best result. The comparative analysis of the summary graph (Table \ref{comparison-symmary-graph}) shows that the proposed method achieves the best scores for all three evaluation criteria for both datasets except for the FDR of DAG-GNN method. For synthetic dataset-1, DAG-GNN method yields a lower score in terms of FDR whereas its SHD is close to the proposed model. Since, FDR only relies on the number of predicted edges, the $(0.00^*)$ FDR score of DAG-GNN cannot be considered a perfect score as the few edges predicted by the method align with the true causal graph but it missed predicting the remaining edges (Figure \ref{true-graph}a).


\begin{table}[t!]
\caption{{Comparison of the predicted temporal full causal graph by different methods for synthetic datasets.}}
\label{comparison-full-graph}
\begin{center}
\begin{small}
\begin{sc}
\setlength\tabcolsep{3pt}
\begin{tabular}{l|ccc|ccc}
\toprule
Method & \multicolumn{3}{c|}{Dataset-1} & \multicolumn{3}{c}{Dataset-2}  \\
  & SHD & F1 & FDR & SHD & F1 & FDR  \\
\midrule
PCMCI \cite{pcmci}   & 69 & 0.16 & 0.90 & 22 & 0.38 & 0.74 \\
PCMCI+ \cite{pcmci+} & 49 & 0.16 & 0.90 & 11 & \textbf{0.62} & 0.55 \\
NOTEARS-MLP \cite{notears-mlp}  & 13 & \underline{0.58} & \underline{0.59} & 20 & 0.44 & 0.70 \\
NTS-NOTEARS  \cite{nts-notears}  & \underline{12} & 0.45 & 0.61 & \underline{8} & 0.33 & \underline{0.33} \\
DAG-GNN  \cite{yu2019dag}    & \underline{12} & 0.00 & 1.00 & 18 & 0.28 & 0.80 \\
DYNOTEARS \cite{dynotears}  & 13 & 0.51 & 0.61 & 18 & 0.25 & 0.80 \\
Proposed      & \textbf{8} & \textbf{0.60} & \textbf{0.45} & \textbf{7} & \underline{0.46} & \textbf{0.25} \\
\bottomrule
\end{tabular}
\end{sc}
\end{small}
\end{center}
\vskip -0.1in
\end{table}

To compare the results of the predicted full causal graph, we considered both the directed edges and the time lag of the edge. From Table \ref{comparison-full-graph}, we can see that our proposed method got the best F1 score and the second-lowest SHD and FDR for dataset-1. For dataset-2, the proposed method yields the best results for all three quality measures. For dataset-1, NTS-NOTEARS generated the target causal graph with smaller SHD and lower F1 scores compared to the proposed method. This can be interpreted as the number of edges generated by NTS-NOTEARS is lower than the number of edges in the true causal graph (low F1 score) and also the 0.5 FDR value means that half of the predicted edges are wrong. Similarly, zero F1 score of DAG-GNN method means that it was not able to find any causal links from the true causal graph. From observing the evaluation results of the full causal graph, we can find that the proposed model is capable of generating fewer false causal links compared to other baseline models.


\begin{table}[t!]
\caption{{Comparison of the predicted summary causal graph by different methods for synthetic datasets.}}
\label{comparison-symmary-graph}
\begin{center}
\begin{small}
\begin{sc}
\setlength\tabcolsep{3pt}
\begin{tabular}{l|ccc|ccc}
\toprule
Method & \multicolumn{3}{c|}{Dataset-1} & \multicolumn{3}{c}{Dataset-2}  \\
  & SHD & F1 & FDR & SHD & F1 & FDR  \\
\midrule
PCMCI \cite{pcmci}   & 10 & 0.54 & 0.62 & 7 & \underline{0.63} & 0.53 \\
PCMCI+ \cite{pcmci+} & 10 & 0.54 & 0.62 & 6 & \textbf{0.66} & 0.50 \\
NOTEARS-MLP \cite{notears-mlp}  & 6 & 0.25 & 0.50 & \textbf{4} & 0.50 & \textbf{0} \\
NTS-NOTEARS  \cite{nts-notears}  & \textbf{3} & \textbf{0.76} & \textbf{0.28} & \underline{5} & 0.44 & 0.33 \\
DAG-GNN  \cite{yu2019dag}    & 5 & 0.44 & \underline{0.33} & 8 & 0.50 & 0.60 \\
DYNOTEARS \cite{dynotears}  & \underline{4} & \underline{0.74} & 0.40 & 7 & 0.36 & 0.60 \\
Proposed      & \textbf{3} & \textbf{0.76} & \textbf{0.28} & \textbf{4} & 0.60 & \underline{0.25} \\
\bottomrule
\end{tabular}
\end{sc}
\end{small}
\end{center}
\vskip -0.1in
\end{table}

\subsection{Qualitative Results}
\label{Qualitative-Results}
\begin{figure}[h!]
  \centering
  \includegraphics[width=\linewidth]{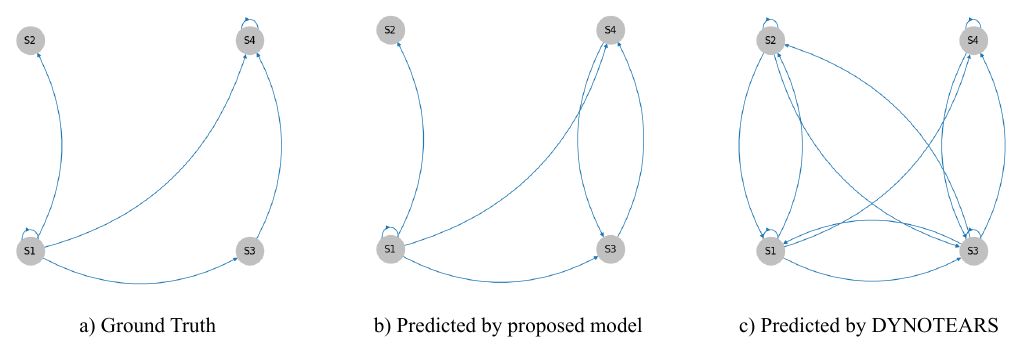}
  \caption{Comparison of ground truth summary causal graph with predicted graphs from DYNOTEARS and our proposed model for synthetic dataset 1.}
  \label{summary-plot-comparison}
\end{figure}

To gain more insights into the discovered causal relationships, we conducted a qualitative analysis of the predicted causal graphs from synthetic dataset-1. Figure \ref{summary-plot-comparison} illustrates the ground truth and predicted causal graphs of the proposed and DYNOTEARS methods using direct graphs. The nodes in the graph model the variables in the dataset and each edge means a causal relationship between the nodes. The self-loop in the causal graph represents that the previous timestep of the variable has a lagged effect on its present timestep. From the predicted causal graphs, we can see that the proposed model predicted the same number of edges to the ground truth graph but failed to predict edge $S4 \to S4$ and estimated a false edge $S3 \to S4$. On the other hand, DYNOTEARS method failed to distinguish the causal effects as mentioned in the ground truth graph. Instead, DYNOTEARS method predicts bidirectional effects between the variables $(S1 \leftrightarrow S2)$,$(S1 \leftrightarrow S3)$,$(S2 \leftrightarrow S3)$, and $(S3 \leftrightarrow S4)$. Whereas $S1$ is the cause variable for the other three variables and there is no causal link between variable $S2$ and $S3$.     


\begin{figure}[h!]
  \centering
  \includegraphics[width=0.8\linewidth]{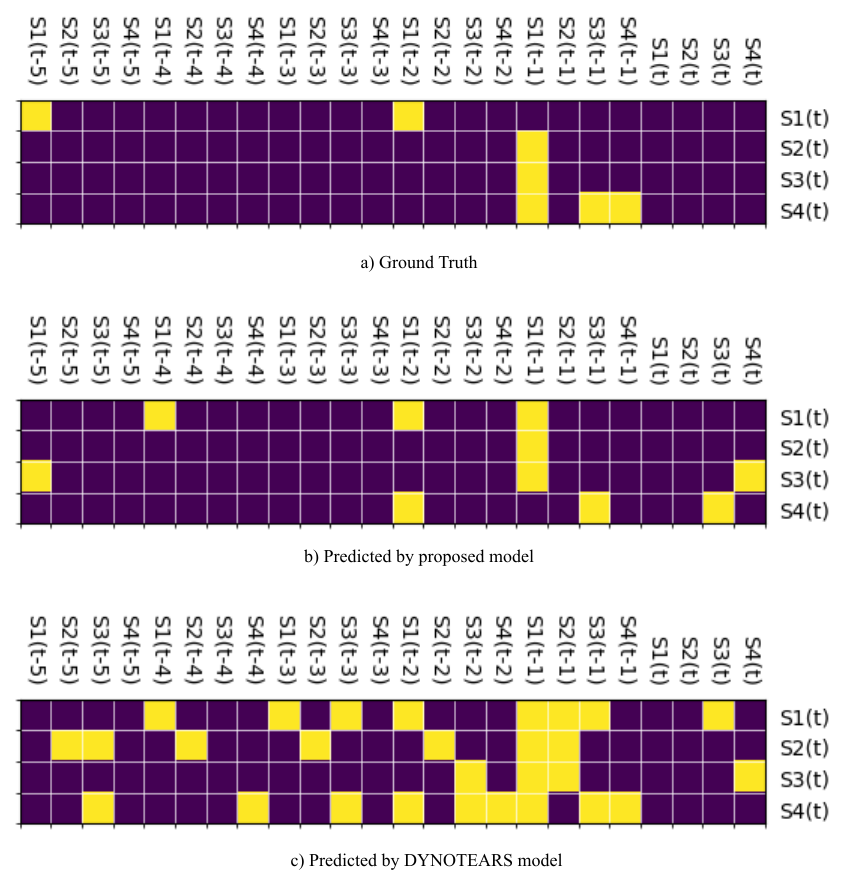}
  \caption{Visualization of ground truth full temporal causal graph of synthetic dataset 1 and the predicted graphs from DYNOTEARS and our proposed model. (The yellow cell means there is a directed edge from the column index to the row index.)}
  \label{full-plot-comparison}
\end{figure}

The full temporal causal graph of the data generation process of synthetic dataset 1 is visualized in the 1st plot of Figure \ref{full-plot-comparison}. Here columns represent the parent variables with the time lag and rows represent the child/target variables. The adjacency matrix of the true graph is sparse which helps us to evaluate the performance of time series causal discovery methods for sparse causal graphs. From the predicted adjacency matrix of the proposed model, we can see that the causal links $\{S1(t-2) \to S1(t), S1(t-1) \to S3(t), S1(t-1) \to S2(t), S3(t-1) \to S4(t)\}$ are predicted correctly. However, the proposed model failed to predict the other three true causal links of the dataset. On the other hand, DYNOTEARS method predicted all edges from the true causal graph except ($S1(t-5) \to S1(t)$) and predicted more false edges than the proposed model. The qualitative and quantitative analysis of the predicted results gives us the intuition that most state-of-the-art models can work better on predicting dense causal graphs than sparse causal graphs like the one used in this evaluation process.


\subsection{Results on Real World Data}
\label{result-real-world}
We applied the proposed and baseline models to the TKE and Arctic Sea Ice datasets to generate causal graphs with time lags 5 and 12, referring to the very fast time evolution of planetary boundary layer turbulence in TKE data and annual seasonality in sea ice data, respectively. The evaluation results of the summary graphs generated by baseline models are summarised in Table \ref{comparison-symmary-graph-real-world}. For the TKE dataset, the proposed model performed similarly to PCMCI+ method. Although NOTEARS-MLP has a lower SHD score, its F1 score is lower than the proposed model. For the Arctic Sea Ice dataset, our proposed method achieves the best F1 and FDR scores. Although the SHD scores of a few baseline models were lower than the proposed model, their F1 and FDR values were not as good.

\begin{table}[t!]
\caption{Comparison of the predicted summary causal graph by different methods for real world datasets.}
\label{comparison-symmary-graph-real-world}
\begin{center}
\begin{small}
\begin{sc}
\setlength\tabcolsep{3pt}
\begin{tabular}{l|ccc|ccc}
\toprule
Method & \multicolumn{3}{c|}{TKE} & \multicolumn{3}{c}{Arctic Sea Ice}  \\
  & SHD & F1 & FDR & SHD & F1 & FDR  \\
\midrule
PCMCI \cite{pcmci}    & 9 & 0.18 & 0.87 & 62 & 0.31 & 0.68 \\
PCMCI+ \cite{pcmci+} & \underline{5} & \textbf{0.44} & \textbf{0.66} & \textbf{50} & 0.32 & \underline{0.57} \\
NOTEARS-MLP \cite{notears-mlp}   & \textbf{4} & 0.33 & \textbf{0.66} & 71 & \underline{0.38} & 0.68 \\
NTS-NOTEARS \cite{nts-notears}   & 6 & \underline{0.40} & 0.71 & \underline{53} & 0.10 & 0.76 \\
DAG-GNN  \cite{yu2019dag}    & 7 & 0.22 & 0.83 & 66 & 0.21 & 0.76 \\
DYNOTEARS \cite{dynotears}  & 8 & 0.00 & 1.00 & 65 & 0.21 & 0.75 \\
Proposed    & \underline{5} & \textbf{0.44} & \textbf{0.66} & 54 & \textbf{0.49} & \textbf{0.56} \\
\bottomrule
\end{tabular}
\end{sc}
\end{small}
\end{center}
\vskip -0.1in
\end{table}

\begin{figure}[h!]
  \centering
  \includegraphics[width=\linewidth]{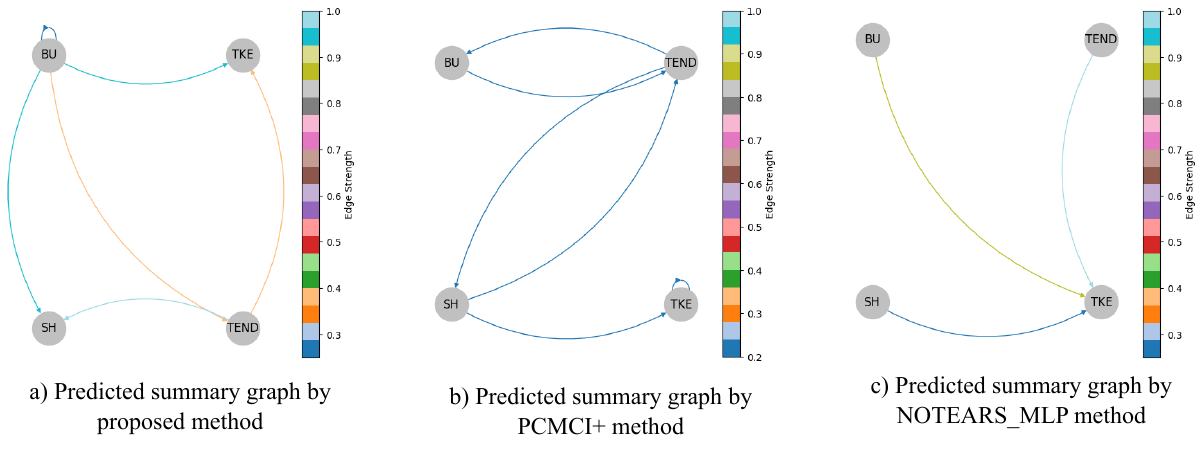}
  \caption{Visualization of predicted summary causal graphs of the TKE dataset using PCMCI+, NOTEARS-MLP, and our proposed model.}
  \label{tke-summary-comparison}
\end{figure}

Due to space limitation, here we only demonstrate the predicted causal summary graphs by NTS-NOTEARS and the proposed method from the TKE dataset. For causal graphs generated by baseline methods on real world datasets, refer to Appendices \ref{tke-all} and \ref{sie-all}. We analyze the predicted causal graphs by comparing the domain graph represented in Figure~\ref{true-graph}. From the predicted graphs in Figure~\ref{tke-summary-comparison}, we can see that the proposed method recovered the correct edges from $BU \to TEND$, $TEND \to TKE$, and the connection between each pair of nodes is unidirectional. On the other hand, PCMCI+ method predicted bidirectional edges for node pairs $BU \to TEND$, and  $SH \to TEND$ and missed the important edge from $TEND$ to $TKE$. On the other hand, NOTEARS-MLP predicted one correct edge $TEND \to TKE$ though it has a lower SHD score. The time lag 5 is used for discovering lagged causal relationships for each variable. The predicted full causal graph can give us an idea about the lagged effect of different variables in this dataset which is illustrated in Appendix \ref{tke-all}. By observing the predicted full and summary graph we see that the proposed model can better identify the lagged and contemporaneous causal links.



\subsection{Ablation Study}

To verify the effectiveness of the proposed model for temporal causal discovery, a comparative study of this model with its variant is shown here. The quantitative results of the generated summary causal graphs of these models are illustrated in Table \ref{ablation-results}. For Causal Conv1D, we used 1D variant of the proposed custom Causal Conv2D layer. To incorporate this Conv1D layer into the model, we flattened the latent representation of the earlier convolution layers of the model. The same training and optimization process was utilized for both models. The ablation study results show that the Custom Conv2D layer improves the causal graphs learning performance of the proposed model with a significant margin for each evaluation score. As we are studying multivariate time series data convolution 2D layers can better learn the inherent features of the data generation than convolution 1D layers.   


\begin{table}[t]
\caption{{Performance Comparison between Custom Causal Conv2D and Conv1D models.}}
\label{ablation-results}
\begin{center}
\begin{small}
\begin{sc}
\setlength\tabcolsep{3pt}
\begin{tabular}{l|ccc|ccc}
\toprule
Dataset & \multicolumn{3}{c|}{Causal Conv1D} & \multicolumn{3}{c}{Causal Conv2D}  \\
  & SHD & F1 & FDR & SHD & F1 & FDR  \\
\midrule
Synthetic Datase-1   & 7 & 0.46 & 0.57 & 3 & 0.77 & 0.28 \\
Synthetic Datase-2 & 8 & 0.33 & 0.66 & 4 & 0.60 & 0.25 \\
TKE    & 6 & 0.40 & 0.71 & 5 & 0.44 & 0.66 \\
Arctic Sea Ice  & 63 & 0.42 & 0.63 & 54 & 0.49 & 0.56 \\
\bottomrule
\end{tabular}
\end{sc}
\end{small}
\end{center}
\vskip -0.1in
\end{table}

\subsection{Robustness Analysis}
To investigate the performance stability of the proposed model, we considered different features of datasets including signal-to-noise ratio (SNR), noise distribution, non-stationary and stationary variable combinations. We generated three versions of the synthetic dataset-1 (Synthetic-1.1 with lower SNR, Synthetic-1.2 with medium SNR, and Synthetic-1.3 with higher SNR comparatively) to evaluate the response of the proposed model to the different SNRs and applied all baseline models to them. The evaluation results of the proposed model for these synthetic datasets are summarized in Table \ref{tbl:comparison-snr-syn} and more results for baseline models are reported in Appendix \ref{SNR}. By observing evaluation results we can see that the performance of the proposed model increases with the increase of SNRs. The non-stationarity property analysis of the TKE and Arctic Sea Ice data revealed that all variables of the TKE data are non-stationary and there is a good mixture of non-stationary and stationary variables in the Arctic Sea Ice data (Test scores available in Appendix \ref{non-st-test}). Our proposed model demonstrated good performance for both of these real world non-stationarity datasets (Section \ref{result-real-world}). The simulated datasets qualify the impact of various noise distributions on the proposed model's performance. For synthetic datasets 1 and 2, respectively, we employed Gaussian and Poisson noise distributions. The evaluation findings confirm that our model can function with both types of noise (Sections \ref{Qualitative-Results} and \ref{Quantitative-Results}) and is not restricted to only these two types of noise. The comprehensive experiments with different cases and the evaluation results confirm the robustness of the proposed model for causal discovery for time series data with diverse properties.

\begin{table}[t]
\caption{Evaluation of the predicted causal graphs by the proposed method for different SNRs.}
\label{tbl:comparison-snr-syn}
\begin{center}
\begin{small}
\begin{sc}
\setlength\tabcolsep{3pt}
\begin{tabular}{l|ccc|ccc}
\toprule
Dataset & \multicolumn{3}{c|}{Summary Graph} & \multicolumn{3}{c}{Full Graph}  \\
  & SHD & F1 & FDR & SHD & F1 & FDR  \\
\midrule
Synthetic-1.1    & 5 & 0.54 & 0.40 & 12 & 0.33 & 0.72 \\
Synthetic-1.2 & 4 & 0.66 & 0.33 & 11 & 0.42 & 0.66 \\
Synthetic-1.3  & 2 & 0.83 & 0.16 & 9 & 0.47 & 0.60 \\
\bottomrule
\end{tabular}
\end{sc}
\end{small}
\end{center}
\vskip -0.1in
\end{table}

\section{Related Works}
\label{related_works}

\noindent\textbf{Causal Convolutional Neural Network (CausalCNN).} Recently, some studies have combined the similarities in hierarchical structures within convolutional neural networks (CNNs) and causal graphs to develop causal networks using CNNs to serve a variety of purposes. \cite{Debbi2021Causal} proposed CexCNN - an explanation technique for CNNs using Pearl’s theory of counterfactual reasoning to improve predictability. \cite{Li2019Knowledge} proposed a Knowledge-oriented Convolutional Neural Network (K-CNN) for causal relation extraction in texts for natural language processing tasks. \cite{Karamjit2017Deep} use attribute pairs to pre-train CNN models for determining pair-wise causal direction in data. \cite{SingH2019Recurrent} proposed a novel Recurrent Convolutional Network (RCN) where they employ recurrence to 3D CNN for causal processing of videos. Earlier, \cite{nauta2019causal} proposed an attention-based CNN approach for time series causal discovery, however, their approach had several limitations including the stationarity assumption in data. It is important to note that our proposed approach is distinguishable from these methods in both functionality and usage.

\noindent\textbf{Causal Discovery on Nonlinear and Non-Stationary Data.} Statistical causal discovery methods, like Granger Causality (GC), cannot handle non-linearity or non-stationarity in data. While some methods extend the traditional causal discovery method to handle non-linearity \cite{gerhardus2020high,bahadori2012causality}, some approaches utilize neural network-for these extensions~\cite{tank2021neural,absar2023neural}. Further, some recent research has made inroads to propose techniques applicable to non-stationary time-series. \cite{cd-nod} proposed Constraint-based causal Discovery from NOnstationary Data (CD-NOD). Introducing a non-parametric principled framework, they demonstrated the efficiency of their approach in forecasting non-stationary data. \cite{Carles2021Causal} proposed a probabilistic deep learning approach, State-Dependent Causal Inference (SDCI), to perform causal discovery in conditionally stationary time-series data. \cite{Wu2022Nonlinear} introduced a functional causal model (FCM) based approach for causal learning in non-stationary data with general nonlinear relationships. \cite{Ferdous2023CDANs} proposed a constraint-based causal discovery approach for autocorrelated and non-stationary time series data (CDANs) to identify lagged and instantaneous causal relationships in autocorrelated and non-stationary time series data. Most recently, \cite{Fujiwara23a} combined Linear Non-Gaussian Acyclic Model (LiNGAM) and the Just-In-Time (JIT) framework to propose a new causal discovery method JIT-LiNGAM to identify causal relations in non-linear and non-stationary data. Recent work also includes deep learning-based methods which infer the causal graph directly from observational data using transformers, generative modeling, and adversarial learning techniques \cite{nauta2019causal,kalainathan2018sam,yu2019dag}, overcoming the non-linearity constraint whereas the challenge of handling non-stationarity still prevails in them. 

\noindent\textbf{Limitations of Related Works.} While these existing methodologies have significantly contributed to the field of time series causal discovery, several challenges persist. The inability to handle high-dimensional non-stationary datasets, scalability issues, and difficulty in distinguishing causation from correlation remain active research areas. Additionally, many methods struggle with interpreting and providing actionable insights, especially in complex real world applications such as climate science.
By addressing the challenges posed by traditional and machine learning based approaches, our method aims to provide a comprehensive solution for effective time series causal discovery.

\section{Conclusion}
\label{conclusion}
We proposed TS-CausalNN, a score-based causal structure learning method for non-linear and non-stationary time series data using custom 2D causal convolution layer. The proposed method utilizes the power of the 2D neural networks to learn the contemporaneous and time-lagged causal relationships of all temporal variables simultaneously. The theory and computational procedure of the method are established from the continuous formulation of the acyclicity constraint of the causal DAG adapted from Zheng et al. \cite{zheng2018dags}. The parallel causal convolution layer block of the proposed model keeps the causal contributor of each target/effect variable segregated from other target variables, which gives better causal stability to the generated DAG structure. The proposed model is very simple and user-friendly as any prior knowledge of variable independence or the data generation model is not required and also there is no assumption regarding error distribution. We conducted extensive experiments on synthetic and real world complex time series datasets to demonstrate the performance of the proposed causal discovery model. Based on the empirical evaluation results, the proposed model demonstrates superior causal graph learning capability compared to the baseline methods. 



The code is available at \href{https://anonymous.4open.science/r/TS-CausalNN-Learning-Temporal-Causal-Relations-from-Non-linear-Non-stationary-Time-Series-Data-41B3}{http://tinyurl.com/TS-CausalNN}.

\section*{Acknowledgment}
This work was partially supported by the DOE Office of Science Early Career Research Program. This work was performed under the auspices of the U.S. Department of Energy (DOE) by LLNL under contract DE-AC52-07NA27344. LLNL-CONF-846980. Faruque, Ali and Wang were also partially supported by grants OAC-1942714 and OAC-2118285 from the U.S. National Science Foundation (NSF).



\bibliographystyle{splncs04}
\bibliography{sample-base}

\begin{thebibliography}{10}
\providecommand{\url}[1]{\texttt{#1}}
\providecommand{\urlprefix}{URL }
\providecommand{\doi}[1]{https://doi.org/#1}

\bibitem{absar2023neural}
Absar, S., Wu, Y., Zhang, L.: Neural time-invariant causal discovery from time series data. In: 2023 International Joint Conference on Neural Networks (IJCNN). pp.~1--8. IEEE (2023)

\bibitem{economics}
Appiah, M.O.: Investigating the multivariate granger causality between energy consumption, economic growth and co2 emissions in ghana. Energy Policy  \textbf{112},  198--208 (2018)

\bibitem{bahadori2012causality}
Bahadori, M.T., Liu, Y.: On causality inference in time series. In: 2012 AAAI Fall Symposium Series (2012)

\bibitem{Carles2021Causal}
{Carles Balsells Rodas}, {Ruibo Tu}, {H. Kjellström}: Causal discovery from conditionally stationary time-series. arXiv.org  (2021)

\bibitem{cohen2014recent}
Cohen, J., Screen, J.A., Furtado, J.C., Barlow, M., Whittleston, D., Coumou, D., Francis, J., Dethloff, K., Entekhabi, D., Overland, J., et~al.: Recent arctic amplification and extreme mid-latitude weather. Nature geoscience  \textbf{7}(9),  627--637 (2014)

\bibitem{Debbi2021Causal}
Debbi, H.: Causal {Explanation} of {Convolutional} {Neural} {Networks}, pp. 633--649. Springer International Publishing (2021)

\bibitem{endo2015racoro}
Endo, S., Fridlind, A.M., Lin, W., Vogelmann, A.M., Toto, T., Ackerman, A.S., McFarquhar, G.M., Jackson, R.C., Jonsson, H.H., Liu, Y.: Racoro continental boundary layer cloud investigations: 2. large-eddy simulations of cumulus clouds and evaluation with in situ and ground-based observations. Journal of Geophysical Research: Atmospheres  \textbf{120}(12),  5993--6014 (2015)

\bibitem{ts-fci}
Entner, D., Hoyer, P.O.: On causal discovery from time series data using fci. Probabilistic graphical models pp. 121--128 (2010)

\bibitem{Ferdous2023CDANs}
Ferdous, M.H., Hasan, U., Gani, M.O.: Cdans: Temporal {Causal} {Discovery} from {Autocorrelated} and {Non}-{Stationary} {Time} {Series} {Data}  (2023)

\bibitem{Fujiwara23a}
Fujiwara, D., Koyama, K., Kiritoshi, K., Okawachi, T., Izumitani, T., Shimizu, S.: Causal discovery for non-stationary non-linear time series data using just-in-time modeling. In: van~der Schaar, M., Zhang, C., Janzing, D. (eds.) Proceedings of the Second Conference on Causal Learning and Reasoning. Proceedings of Machine Learning Research, vol.~213, pp. 880--894. PMLR (11--14 Apr 2023), \url{https://proceedings.mlr.press/v213/fujiwara23a.html}

\bibitem{lpcmci}
Gerhardus, A., Runge, J.: High-recall causal discovery for autocorrelated time series with latent confounders. Advances in Neural Information Processing Systems  \textbf{33},  12615--12625 (2020)

\bibitem{gerhardus2020high}
Gerhardus, A., Runge, J.: High-recall causal discovery for autocorrelated time series with latent confounders. Advances in Neural Information Processing Systems  \textbf{33},  12615--12625 (2020)

\bibitem{gu2018recent}
Gu, J., Wang, Z., Kuen, J., Ma, L., Shahroudy, A., Shuai, B., Liu, T., Wang, X., Wang, G., Cai, J., et~al.: Recent advances in convolutional neural networks. Pattern recognition  \textbf{77},  354--377 (2018)

\bibitem{les-arm}
Gustafson, W.I., Vogelmann, A.M., Li, Z., Cheng, X., Dumas, K.K., Endo, S., Johnson, K.L., Krishna, B., Fairless, T., Xiao, H.: The large-eddy simulation (les) atmospheric radiation measurement (arm) symbiotic simulation and observation (lasso) activity for continental shallow convection. Bulletin of the American Meteorological Society  \textbf{101}(4),  E462--E479 (2020)

\bibitem{hasan2023survey}
Hasan, U., Hossain, E., Gani, M.O.: A survey on causal discovery methods for iid and time series data. Transactions on Machine Learning Research  (2023)

\bibitem{medicine}
Heckerman, D.E., Horvitz, E.J., Nathwani, B.N.: Toward normative expert systems: Part i the pathfinder project. Methods of information in medicine  \textbf{31}(02),  90--105 (1992)

\bibitem{tke}
Hinze, J.O.: Turbulence. McGraw-Hill (1975)

\bibitem{cd-nod}
Huang, B., Zhang, K., Zhang, J., Ramsey, J., Sanchez-Romero, R., Glymour, C., Sch{\"o}lkopf, B.: Causal discovery from heterogeneous/nonstationary data. The Journal of Machine Learning Research  \textbf{21}(1),  3482--3534 (2020)

\bibitem{huang2021benchmarking}
Huang, Y., Kleindessner, M., Munishkin, A., Varshney, D., Guo, P., Wang, J.: Benchmarking of data-driven causality discovery approaches in the interactions of arctic sea ice and atmosphere. Frontiers in big Data  \textbf{4},  642182 (2021)

\bibitem{kalainathan2018sam}
Kalainathan, D., Goudet, O., Guyon, I., Lopez-Paz, D., Sebag, M.: Sam: Structural agnostic model, causal discovery and penalized adversarial learning  (2018)

\bibitem{Karamjit2017Deep}
{Karamjit Singh}, {Garima Gupta}, {L. Vig}, {Gautam M. Shroff}, {P. Agarwal}: Deep {Convolutional} {Neural} {Networks} for {Pairwise} {Causality}. arXiv.org  (2017)

\bibitem{ml-application}
Koller, D., Friedman, N.: Probabilistic graphical models: principles and techniques. MIT press (2009)

\bibitem{Li2019Knowledge}
Li, P., Mao, K.: Knowledge-oriented convolutional neural network for causal relation extraction from natural language texts. Expert Systems with Applications  \textbf{115},  512--523 (1 2019)

\bibitem{luo2018changes}
Luo, D., Chen, X., Dai, A., Simmonds, I.: Changes in atmospheric blocking circulations linked with winter arctic warming: A new perspective. Journal of Climate  \textbf{31}(18),  7661--7678 (2018)

\bibitem{luo2019weakened}
Luo, D., Chen, X., Overland, J., Simmonds, I., Wu, Y., Zhang, P.: Weakened potential vorticity barrier linked to recent winter arctic sea ice loss and midlatitude cold extremes. Journal of Climate  \textbf{32}(14),  4235--4261 (2019)

\bibitem{nauta2019causal}
Nauta, M., Bucur, D., Seifert, C.: Causal {Discovery} with {Attention}-{Based} {Convolutional} {Neural} {Networks}. Machine Learning and Knowledge Extraction  \textbf{1}(1),  312--340 (jan 7 2019)

\bibitem{dynotears}
Pamfil, R., Sriwattanaworachai, N., Desai, S., Pilgerstorfer, P., Georgatzis, K., Beaumont, P., Aragam, B.: Dynotears: Structure learning from time-series data. In: International Conference on Artificial Intelligence and Statistics. pp. 1595--1605. PMLR (2020)

\bibitem{Pearl1991ProbabilisticRI}
Pearl, J.: Probabilistic reasoning in intelligent systems - networks of plausible inference. In: Morgan Kaufmann series in representation and reasoning (1991), \url{https://api.semanticscholar.org/CorpusID:32583695}

\bibitem{pearl2000models}
Pearl, J.: Models, reasoning and inference. Cambridge, UK: CambridgeUniversityPress  \textbf{19}(2), ~3 (2000)

\bibitem{peters2017elements}
Peters, J., Janzing, D., Sch{\"o}lkopf, B.: Elements of causal inference: foundations and learning algorithms. The MIT Press, Cambridge, MA, USA (2017)

\bibitem{rajapakse2007learning}
Rajapakse, J.C., Zhou, J.: Learning effective brain connectivity with dynamic bayesian networks. Neuroimage  \textbf{37}(3),  749--760 (2007)

\bibitem{pcmci+}
Runge, J.: Discovering contemporaneous and lagged causal relations in autocorrelated nonlinear time series datasets. In: Peters, J., Sontag, D. (eds.) Proceedings of the 36th Conference on Uncertainty in Artificial Intelligence (UAI). Proceedings of Machine Learning Research, vol.~124, pp. 1388--1397. PMLR (03--06 Aug 2020)

\bibitem{runge2019inferring}
Runge, J., Bathiany, S., Bollt, E., Camps-Valls, G., Coumou, D., Deyle, E., Glymour, C., Kretschmer, M., Mahecha, M.D., Mu{\~n}oz-Mar{\'\i}, J., et~al.: Inferring causation from time series in earth system sciences. Nature communications  \textbf{10}(1), ~2553 (2019)

\bibitem{Runge2019InferringCF}
Runge, J., Bathiany, S., Bollt, E.M., Camps-Valls, G., Coumou, D., Deyle, E.R., Glymour, C., Kretschmer, M., Mahecha, M.D., Mu{\~n}oz-Mar{\'i}, J., van Nes, E.H., Peters, J., Quax, R., Reichstein, M., Scheffer, M., Scholkopf, B., Spirtes, P., Sugihara, G., Sun, J., Zhang, K., Zscheischler, J.: Inferring causation from time series in earth system sciences. Nature Communications  \textbf{10} (2019), \url{https://api.semanticscholar.org/CorpusID:189819550}

\bibitem{pcmci}
Runge, J., Nowack, P., Kretschmer, M., Flaxman, S., Sejdinovic, D.: Detecting and quantifying causal associations in large nonlinear time series datasets. Science Advances  \textbf{5}(11),  eaau4996 (2019). \doi{10.1126/sciadv.aau4996}

\bibitem{finance}
Sanford, A.D., Moosa, I.A.: A bayesian network structure for operational risk modelling in structured finance operations. Journal of the Operational Research Society  \textbf{63},  431--444 (2012)

\bibitem{serreze2015arctic}
Serreze, M.C., Stroeve, J.: Arctic sea ice trends, variability and implications for seasonal ice forecasting. Philosophical Transactions of the Royal Society A: Mathematical, Physical and Engineering Sciences  \textbf{373}(2045),  20140159 (2015)

\bibitem{Shah_2020}
Shah, R.D., Peters, J.: The hardness of conditional independence testing and the generalised covariance measure. The Annals of Statistics  \textbf{48}(3) (Jun 2020). \doi{10.1214/19-aos1857}

\bibitem{simmonds2015comparing}
Simmonds, I.: Comparing and contrasting the behaviour of arctic and antarctic sea ice over the 35 year period 1979-2013. Annals of Glaciology  \textbf{56}(69),  18--28 (2015)

\bibitem{simmonds2014physical}
Simmonds, I., Govekar, P.D.: What are the physical links between arctic sea ice loss and eurasian winter climate? Environmental Research Letters  \textbf{9}(10),  101003 (2014)

\bibitem{SingH2019Recurrent}
SingH, G., Cuzzolin, F.: Recurrent {Convolutions} for {Causal} 3d {CNNs}. In: 2019 {IEEE}/{CVF} {International} {Conference} on {Computer} {Vision} {Workshop} ({ICCVW}). IEEE (10 2019)

\bibitem{wrf_skamarock}
Skamarock, W.C., Klemp, J.B., Dudhia, J., Gill, D.O., Liu, Z., Berner, J., Wang, W., Powers, J., Duda, M., Barker, D., et~al.: A description of the advanced research wrf version 4. NCAR tech. note ncar/tn-556+ str  \textbf{145} (2019)

\bibitem{spirtes2000causation}
Spirtes, P., Glymour, C.N., Scheines, R.: Causation, prediction, and search. MIT press, Cambridge, MA, USA (2000)

\bibitem{nts-notears}
Sun, X., Schulte, O., Liu, G., Poupart, P.: Nts-notears: Learning nonparametric dbns with prior knowledge. In: International Conference on Artificial Intelligence and Statistics. pp. 1942--1964. PMLR (2023)

\bibitem{tank2021neural}
Tank, A., Covert, I., Foti, N., Shojaie, A., Fox, E.B.: Neural granger causality. IEEE Transactions on Pattern Analysis and Machine Intelligence  \textbf{44}(8),  4267--4279 (2021)

\bibitem{Wu2022Nonlinear}
Wu, T., Wu, X., Wang, X., Liu, S., Chen, H.: Nonlinear {Causal} {Discovery} in {Time} {Series}. In: Proceedings of the 31st {ACM} {International} {Conference} on {Information} \&amp; {Knowledge} {Management}. ACM (oct 17 2022)

\bibitem{yao2017increased}
Yao, Y., Luo, D., Dai, A., Simmonds, I.: Increased quasi stationarity and persistence of winter ural blocking and eurasian extreme cold events in response to arctic warming. part i: Insights from observational analyses. Journal of Climate  \textbf{30}(10),  3549--3568 (2017)

\bibitem{yu2019dag}
Yu, Y., Chen, J., Gao, T., Yu, M.: Dag-gnn: Dag structure learning with graph neural networks. In: International Conference on Machine Learning. pp. 7154--7163. PMLR (2019)

\bibitem{zheng2018dags}
Zheng, X., Aragam, B., Ravikumar, P., Xing, E.P.: {DAGs with NO TEARS: Continuous Optimization for Structure Learning}. In: Advances in Neural Information Processing Systems (2018)

\bibitem{notears-mlp}
Zheng, X., Dan, C., Aragam, B., Ravikumar, P., Xing, E.P.: {Learning sparse nonparametric DAGs}. In: International Conference on Artificial Intelligence and Statistics (2020)

\end{thebibliography}


\newpage

\appendix

\centerline{\textbf{\Large APPENDIX}}

\section{Synthetic Dataset 2}
\label{syn-data-2}
The synthetic dataset-2 is generated using the following equations and the noise signals used in this dataset are generated by the Poisson distribution. Here we used the exponential non-linearity in the same causal relationship of the synthetic dataset-1 and excluded the seasonal features by removing the cosine function. Also, the signal-to-noise ratio of this dataset is different from dataset-1.    
\begin{equation}
        S1_t = 0.7 e^{\frac{-S1^2_{t-2}\times S1^2_{t-5}}{2}} + \varepsilon 1 
    \end{equation}
    \begin{equation}
    S2_t = 2 e^{\frac{S1^2_{t-1}}{2}} + 0.5 e^{\frac{S1^2_{t}}{2}} + \varepsilon 2
    \end{equation}
    \begin{equation}
    S3_t = -5.05 e^{\frac{-S1^2_{t-1}}{2}} + \varepsilon 3
    \end{equation}
    \begin{equation}
    S4_t = -1.15 e^{\frac{-S1^2_{t-1}}{2}} + 2.35e^{\frac{-S3^2_{t-1}}{2}} + 1.5 e^{\frac{-S4^2_{t-1}}{2}} + 3 e^{\frac{-S3^2_{t}}{2}} + \varepsilon4
    \end{equation}

\section{Method Hyperparameters}
\label{method-hyper}
To find the best hyperparameters for baseline methods we started using the parameters suggested by the authors and gradually tuned those values to obtain better evaluation results. The best results obtained with tuned hyperparameters are reported in the comparative analysis. The parameters used for generating evaluation results are given here.
\begin{itemize}
  \item PCMCI 
    \begin{itemize}
      \item Conditional Independence Test = ParCorr
      \item $tau\_max$ = Maximum time lag
      \item $pc\_alpha$ = None [So the model will use the optimal value from the list $\{0.05, 0.1, 0.2, 0.3, 0.4, 0.5\}$]
      \item $alpha\_level$ = 0.01
    \end{itemize}
  \item PCMCI+ 
    \begin{itemize}
      \item Conditional Independence Test = ParCorr
      \item $tau\_max$ = Maximum time lag
      \item $pc\_alpha$ = None [So the model will use the optimal value from the list $\{0.001, 0.005, 0.01, 0.025, 0.05\}$]
    \end{itemize}
  \item NOTEARS-MLP
    \begin{itemize}
        \item lambda1 = 0.01
        \item lambda2 = 0.01
        \item rho = 1.0
        \item alpha = 0.0
        \item  $w\_threshold$ = 0.3
        
    \end{itemize}
  \item NTS-NOTEARS
    \begin{itemize}
        \item lambda1 = 0.0005
        \item lambda2 = 0.001
        \item $w\_threshold$ = 0.3
        \item rho = 1.0
        \item alpha = 0.0
        \item $number\_of\_lags$ = Maximum time lag
    \end{itemize}
  \item DYNOTEARS
    \begin{itemize}
        \item $tau\_max$ = Maximum time lag
        \item $w\_threshold$ = 0.01
        \item $lambda\_w$ = 0.05
        \item $lambda\_a$ = 0.05
    \end{itemize}
  \item Proposed Method
    \begin{itemize}
        \item lambda1 = 0.1
        \item alpha=0.0
        \item rho = 1.0
        \item beta = 0.1
        \item gamma = 0.25
        \item $w\_threshold$ = 0.3
    \end{itemize}
\end{itemize}

\section{Robustness tests using data with different Signal-to-Noise ratios }
\label{SNR}
We have generated three versions of the synthetic dataset-1 to analyze the performance of the proposed model for different signal-to-noise ratios (SNRs) of datasets. The SNRs of these synthetic dataset versions are given in Table \ref{snr-data}. From the SNR values, we can see that the Synthetic dataset-3 has a higher signal component concerning the noise. And the synthetic dataset-1 has a comparatively lower signal component.        

\begin{table}[ht!]
\caption{Signal-to-Noise ratio of synthetic datasets.}
\label{snr-data}
\begin{center}
\begin{small}
\begin{sc}
\setlength\tabcolsep{3pt}
\begin{tabular}{l|c|c|c|c|c}
\toprule
 Dataset & S1 & S2 & S3 & S4 & Average \\
 \midrule
Synthetic-1.1 & 0.10  & 0.005 & 0.63 & 0.01  & 0.25 \\
Synthetic-1.2 & 0.19  & 0.001  & 1.93  & 0.06  & 0.55 \\
Synthetic-1.3 & 0.52 & 0.02  & 1.37  & 0.56 & 0.62 \\
\bottomrule
\end{tabular}
\end{sc}
\end{small}
\end{center}
\vskip -0.1in
\end{table}

All the baseline models are applied to these synthetic datasets to quantify the changes in the evaluation performance of the generated causal graphs due to the different noise magnitudes. The evaluation results for the summary and full temporal causal graphs are summarized in Table~\ref{snr-symmary-graph} and Table \ref{snr-full-graph}. The best results are marked in bold text and underlined values represent the second best score. Some baseline models generated empty graphs, so we placed a "NE" for those methods in the comparison table instead of mentioning the zero (0) F1 score and FDR as 1. From Table \ref{snr-symmary-graph} we can see that the proposed method achieved the best SHD for all the datasets. For second and third datasets yielded the best F1 score with the lowest FDR, which means the edges predicted by the proposed model are mostly true.

\begin{table}[t]
\caption{Comparison of the predicted summary causal graph by different methods for synthetic datasets with different SNR.}
\label{snr-symmary-graph}
\begin{center}
\begin{small}
\begin{sc}
\setlength\tabcolsep{3pt}
\begin{tabular}{l|ccc|ccc|ccc}
\toprule
Method & \multicolumn{3}{c|}{Synthtic-1.1} & \multicolumn{3}{c}{Synthtic-1.2} & \multicolumn{3}{c}{Synthtic-1.3}  \\
  & SHD & F1 & FDR & SHD & F1 & FDR & SHD & F1 & FDR \\
\midrule
PCMCI \cite{pcmci}   &10 & 0.54& 0.62 & 10 & \underline{0.54} & 0.62 & 10 & 0.54 & 0.62  \\
PCMCI+ \cite{pcmci+} & 8 & \textbf{0.60} & 0.57 & 10 & \underline{0.54} & 0.62 & 9 & 0.54 & 0.62  \\
NOTEARS-MLP \cite{notears-mlp}     & \underline{6} &  0.25 & \underline{0.50} & 
 8 & 0.20 & 0.75           & 6 & 0.25 & 0.50  \\
NTS-NOTEARS  \cite{nts-notears}  & ne & ne & ne & ne & ne & ne             & 6 & 0.25 & 0.50  \\
DAG-GNN  \cite{yu2019dag}    & ne & ne & ne & ne & ne & ne                & \underline{3} & \underline{0.66} & \textbf{0.00}  \\
DYNOTEARS \cite{dynotears}  & 7 & 0.45 & 0.57 & \underline{7} & 0.45 & \underline{0.57}                 & 7 & 0.63 & 0.53  \\
Proposed     & \textbf{5} & \underline{0.54} & \textbf{0.40} & \textbf{4} & \textbf{0.66} & \textbf{0.33}   & \textbf{2} & \textbf{0.83} & \underline{0.16}  \\
\bottomrule
\end{tabular}
\end{sc}
\end{small}
\end{center}
\vskip -0.1in
\end{table}

The evaluation results of the predicted full temporal causal graphs by all baseline methods for the synthetic datasets are shown in Table \ref{snr-full-graph}. Carefully observing the evaluation result we see that except for the zero (0.00) FDR of DAG-GNN method, the proposed method achieved the best scores in SHD and FDR evaluation criteria for all three datasets. For the first synthetic dataset, the proposed method generated the second best F1 score that is lower than PCMCI+. Overall from the comparative analysis of the evaluation results, we can conclude that the proposed model worked better for different SNRs.

\begin{table}[t]
\caption{Comparison of the predicted full causal graph by different methods for synthetic datasets with different SNR.}
\label{snr-full-graph}
\begin{center}
\begin{small}
\begin{sc}
\setlength\tabcolsep{3pt}
\begin{tabular}{l|ccc|ccc|ccc}
\toprule
Method & \multicolumn{3}{c|}{Synthtic-1.1} & \multicolumn{3}{c}{Synthtic-1.2} & \multicolumn{3}{c}{Synthtic-1.3}  \\
  & SHD & F1 & FDR & SHD & F1 & FDR & SHD & F1 & FDR \\
\midrule
PCMCI \cite{pcmci}   &50 & 0.19& 0.89 & 54 & 0.20 & 0.88 & 61 & 0.18 & 0.89  \\
PCMCI+ \cite{pcmci+} & 46 & 0.20 & 0.88 & 56 & 0.17& 0.90 & 39 & 0.23 & 0.86  \\
NOTEARS-MLP \cite{notears-mlp}     & 26&  0.31 & 0.80 & 23& 0.37 & 0.76           & 27 & 0.30 & 0.81  \\
NTS-NOTEARS  \cite{nts-notears}      & ne & ne & ne & ne & ne & ne & \textbf{7} & 0.25 & \textbf{0.50}  \\
DAG-GNN  \cite{yu2019dag}           & ne & ne & ne & ne & ne & ne  & 10 & 0.00 & 1.00  \\
DYNOTEARS \cite{dynotears}  & 24 & 0.20 & 0.86 & 23 & 0.20 & 0.86                 & 23 & 0.34 & 0.78  \\
Proposed     & \textbf{12} & \textbf{0.33} & \textbf{0.72} & \textbf{11} & \textbf{0.42} & \textbf{0.66}   & 9 & \textbf{0.47} & 0.60  \\
\bottomrule
\end{tabular}
\end{sc}
\end{small}
\end{center}
\vskip -0.1in
\end{table}

\section{Study with datasets with only lagged causal links}

\subsection{Datasets}
To evaluate the performance of our proposed causal discovery method on the

         we have used synthetic datasets. As we know the ground truth causal graph for the synthetic datasets, we can measure and compare the learned causal graph easily. We generate a time series dataset consisting of four non-linear variables using Gaussian white noise $\varepsilon$ (Dataset-1). The mathematical description of each variable is given in Equations 10 to 13. The non-linear characteristic is incorporated in the generation of synthetic dataset to mimic the dynamic properties of real world natural system data. The corresponding true causal graph for the time-series data is given in Figure \ref{true-graph}a.
    \begin{equation}
        S1_t = 2 \{cos(\frac{t}{10}) + log(|S1_{t-2} - S1_{t-5}| + 1)\} + 0.1 \varepsilon 1 
    \end{equation}
    \begin{equation}
    S2_t = 12 e^{\frac{S1^2_{t-1}}{2}} + \varepsilon 2
    \end{equation}
    \begin{equation}
    S3_t = -10.5 e^{\frac{-S1^2_{t-1}}{2}} + \varepsilon 3
    \end{equation}
    \begin{equation}
    S4_t = -11.5 e^{\frac{-S1^2_{t-1}}{2}} + 13.5e^{\frac{-S3^2_{t-1}}{2}} + 1.2 e^{\frac{-S4^2_{t-1}}{2}} + \varepsilon4
    \end{equation}


We have also generated another version of the synthetic dataset (Dataset-2) from the previous causal graph without the cosine part of the data generation formulas. For this case, we used the Poisson distribution to incorporate random noise into the dataset. The goal is to find the performance and robustness of the causal discovery methods for the time series dataset without having wave-shape-like features with non-Gaussian noise. The formula used to generate this dataset is available in Appendix \ref{syn-data-2}.

\subsection{Results}
\label{Quantitative-Results-lagged}
To evaluate the performance of the selected state-of-the-art methods we compared the predicted causal graph in both the summary and full temporal graph setting. The qualitative comparison of the baseline methods is reported in Table \ref{comparison-symmary-graph} and Table \ref{comparison-full-graph}. The best results are marked in bold text and underlined values represent the second best score. To get better prediction results, we have applied each method multiple times on the same dataset and reported the best result. The comparative analysis of the summary graph (Table \ref{comparison-symmary-graph}) shows that the proposed method achieves the best scores for all three evaluation criteria for both datasets except for the FDR of DAG-GNN method. For synthetic dataset-1, DAG-GNN method yields a lower score in terms of FDR whereas its SHD is close to the proposed model. Since, FDR only relies on the number of predicted edges, the $(0.00^*)$ FDR score of DAG-GNN cannot be considered a perfect score as the few edges predicted by the method align with the true causal graph but it missed predicting the remaining edges (Figure \ref{true-graph}a).

\begin{table}[t]
\caption{Comparison of the predicted summary causal graph by different methods for synthetic datasets.}
\label{comparison-symmary-graph-lagged}
\begin{center}
\begin{small}
\begin{sc}
\setlength\tabcolsep{3pt}
\begin{tabular}{l|ccc|ccc}
\toprule
Method & \multicolumn{3}{c|}{Dataset-1} & \multicolumn{3}{c}{Dataset-2}  \\
  & SHD & F1 & FDR & SHD & F1 & FDR  \\
\midrule
PCMCI \cite{pcmci}   & 10 & 0.54 & 0.62 & 9 & 0.57 & 0.60 \\
PCMCI+ \cite{pcmci+} & 9 & 0.54 & 0.62 & 7 & \underline{0.63} & 0.53 \\
NOTEARS-MLP \cite{notears-mlp}  & 6 & 0.25 & 0.50 & \underline{6} & 0.25 & \underline{0.50} \\
NTS-NOTEARS  \cite{nts-notears}  & 6 & 0.25 & 0.50 & 7 & 0.36 & 0.60 \\
DAG-GNN  \cite{yu2019dag}    & \underline{3} & \underline{0.66} & \textbf{0.00$^*$} & 7 & 0.22 & 0.66 \\
DYNOTEARS \cite{dynotears}  & 7 & 0.63 & 0.53 & 7 & 0.53 & 0.55 \\
Proposed      & \textbf{2} &\textbf{0.83} & \underline{0.16} & \textbf{3} & \textbf{0.72} & \textbf{0.20} \\
\bottomrule
\end{tabular}
\end{sc}
\end{small}
\end{center}
\vskip -0.1in
\end{table}

To compare the results of the predicted full causal graph, we considered both the directed edges and the time lag of the edge. From Table \ref{comparison-full-graph}, we can see that our proposed method got the best F1 score and the second-lowest SHD and FDR for dataset-1. For dataset-2, the proposed method yields the best results for all three quality measures. For dataset-1, NTS-NOTEARS generated the target causal graph with smaller SHD and lower F1 scores compared to the proposed method. This can be interpreted as the number of edges generated by NTS-NOTEARS is lower than the number of edges in the true causal graph (low F1 score) and also the 0.5 FDR value means that half of the predicted edges are wrong. Similarly, zero F1 score of DAG-GNN method means that it was not able to find any causal links from the true causal graph. From observing the evaluation results of the full causal graph, we can find that the proposed model is capable of generating fewer false causal links compared to other baseline models.

\begin{table}[t]
\caption{Comparison of the predicted temporal full causal graph by different methods for synthetic datasets.}
\label{comparison-full-graph-lagged}
\begin{center}
\begin{small}
\begin{sc}
\setlength\tabcolsep{3pt}
\begin{tabular}{l|ccc|ccc}
\toprule
Method & \multicolumn{3}{c|}{Dataset-1} & \multicolumn{3}{c}{Dataset-2}  \\
  & SHD & F1 & FDR & SHD & F1 & FDR  \\
\midrule
PCMCI \cite{pcmci}   & 61 & 0.18 & 0.80 & 32 & 0.30 & 0.82 \\
PCMCI+ \cite{pcmci+} & 39 & 0.23 & 0.86 & 23 & \underline{0.37} & 0.76 \\
NOTEARS-MLP \cite{notears-mlp}  & 27 & 0.31 & 0.81 & 24 & 0.33 & 0.77 \\
NTS-NOTEARS  \cite{nts-notears}  & \textbf{7} & 0.22 & \textbf{0.50} & \underline{9} & 0.31 & \underline{0.66} \\
DAG-GNN  \cite{yu2019dag}    & 10 & 0.00 & 1.00 & 10 & 0.00 & 1.00 \\
DYNOTEARS \cite{dynotears}  & 23 & \underline{0.34} & 0.78 & 28 & 0.17 & 0.88 \\
Proposed      & \underline{9} & \textbf{0.47} & \underline{0.60} & \textbf{7} & \textbf{0.46} & \textbf{0.50} \\
\bottomrule
\end{tabular}
\end{sc}
\end{small}
\end{center}
\vskip -0.1in
\end{table}

\section{Non-Stationarity Test Results for Real World Datasets}
\label{non-st-test}
The non-stationarity feature of the real world TKE and Arctic Sea Ice datasets is evaluated using the Augmented Dickey–Fuller test (ADF) and Kwiatkowski-Phillips-Schmidt-Shin test (KPSS) statistical test methods for time series data. The ADF test method assumes a null hypothesis: the time series has a unit root and is not stationary. Then try to reject the null hypothesis and if failed to be rejected, it suggests the time series is not stationarity. For the ADF test, if the p-value of a time series is higher than the 0.05 alpha level the null hypothesis cannot be rejected. So the time series is not stationary. The KPSS test works in a somewhat similar manner to the ADF test but assumes an inverse null hypothesis. The null hypothesis of the KPSS method is that the time series is stationary. If the p-value is less than 0.05 alpha level, we can reject the null hypothesis and derive that the time series is not stationary.

\begin{table}[t]
\caption{Stationarity test results for variables of the TKE dataset.}
\label{stn-tke-data}
\begin{center}
\begin{small}
\begin{sc}
\setlength\tabcolsep{3pt}
\begin{tabular}{l|cc|cc}
\toprule
Variables & \multicolumn{2}{c|}{ADF Test} & \multicolumn{2}{c}{KSSP Test}  \\
  & P-value & Stationary & P-value & Stationary  \\
\midrule
SH   & 0.32 & No & 0.01 & No \\
BU & 0.72 & No & 0.01 &No \\
TEND  & 0.66 & No & 0.01 & No \\
TKE & 0.58 & No & 0.01 & No \\
\bottomrule
\end{tabular}
\end{sc}
\end{small}
\end{center}
\vskip -0.1in
\end{table}

The statistical non-stationarity test results for the TKE dataset are given in Table \ref{stn-tke-data} and the results for the Arctic Sea Ice data are available in Table \ref{stn-ice-data}. The non-stationarity test results revealed that the TKE dataset contains only non-stationary variables and both test methods have agreement on the test outcome. In the Arctic Sea Ice dataset, the ADF test found 4 non-stationary variables and the KSSP method found 3 non-stationary variables. Therefore we can say that the Arctic Sea Ice data have a mixture of both non-stationary and stationary variables.

\begin{table}[t]
\caption{Stationarity test results for variables of the Arctic Sea Ice dataset.}
\label{stn-ice-data}
\begin{center}
\begin{small}
\begin{sc}
\setlength\tabcolsep{3pt}
\begin{tabular}{l|cc|cc}
\toprule
Variables & \multicolumn{2}{c|}{ADF Test} & \multicolumn{2}{c}{KSSP Test}  \\
  & P-value & Stationary & P-value & Stationary  \\
\midrule
HFLX   & 0.10 & \textbf{No} & 0.07 & \textbf{Yes} \\
CC & 0.00 & \textbf{Yes}  & 0.02 & \textbf{No} \\
SW  & 0.00 & Yes & 0.10 & Yes \\
U10m & 0.00 & Yes & 0.10 & Yes \\
SLP   & 0.00 & Yes & 0.10 & Yes \\
PRE & 0.00 & Yes & 0.09 & Yes \\
ICE  & 0.75 &  \textbf{No} & 0.01 &  \textbf{No} \\
LW & 0.28 &  \textbf{No} & 0.10 & \textbf{Yes}  \\
V10m   & 0.00 & Yes & 0.10 & Yes \\
CW & 0.00 & Yes & 0.10 & Yes \\
GH  & 0.01 & Yes & 0.10 & Yes \\
RH & 0.11 &  \textbf{No} & 0.01 &  \textbf{No} \\
\bottomrule
\end{tabular}
\end{sc}
\end{small}
\end{center}
\vskip -0.1in
\end{table}

\begin{table}[t]
\caption{Variables in the Arctic Sea Ice Data}
\label{tab:ice-data}
\begin{center}
\begin{small}
\begin{sc}
\setlength\tabcolsep{3pt}
\begin{tabular}{cc}
\toprule
Variable Abbreviation & Full Name  \\
\midrule
HFLX  & Heat Flux\\
CC & Cloud Cover\\
SW  & Net Shortwave Flux\\
U10m &  Zonal wind at 10m\\
SLP   &  Sea level pressure\\
PRE & Total Precipitation\\
ICE  & Sea Ice\\
LW & Net Longwave Flux\\
V10m   & Meridional Wind at 10m\\
CW & Total Cloud Wate Path\\
GH  & Geopotential Height\\
RH & Relative Humidity\\
\bottomrule
\end{tabular}
\end{sc}
\end{small}
\end{center}
\vskip -0.1in
\end{table}

\section{More Results of the TKE Data}
\label{tke-all}
\begin{figure*}[ht]
\vskip 0.2in
\begin{center}
\centerline{\includegraphics[width=0.8\textwidth]{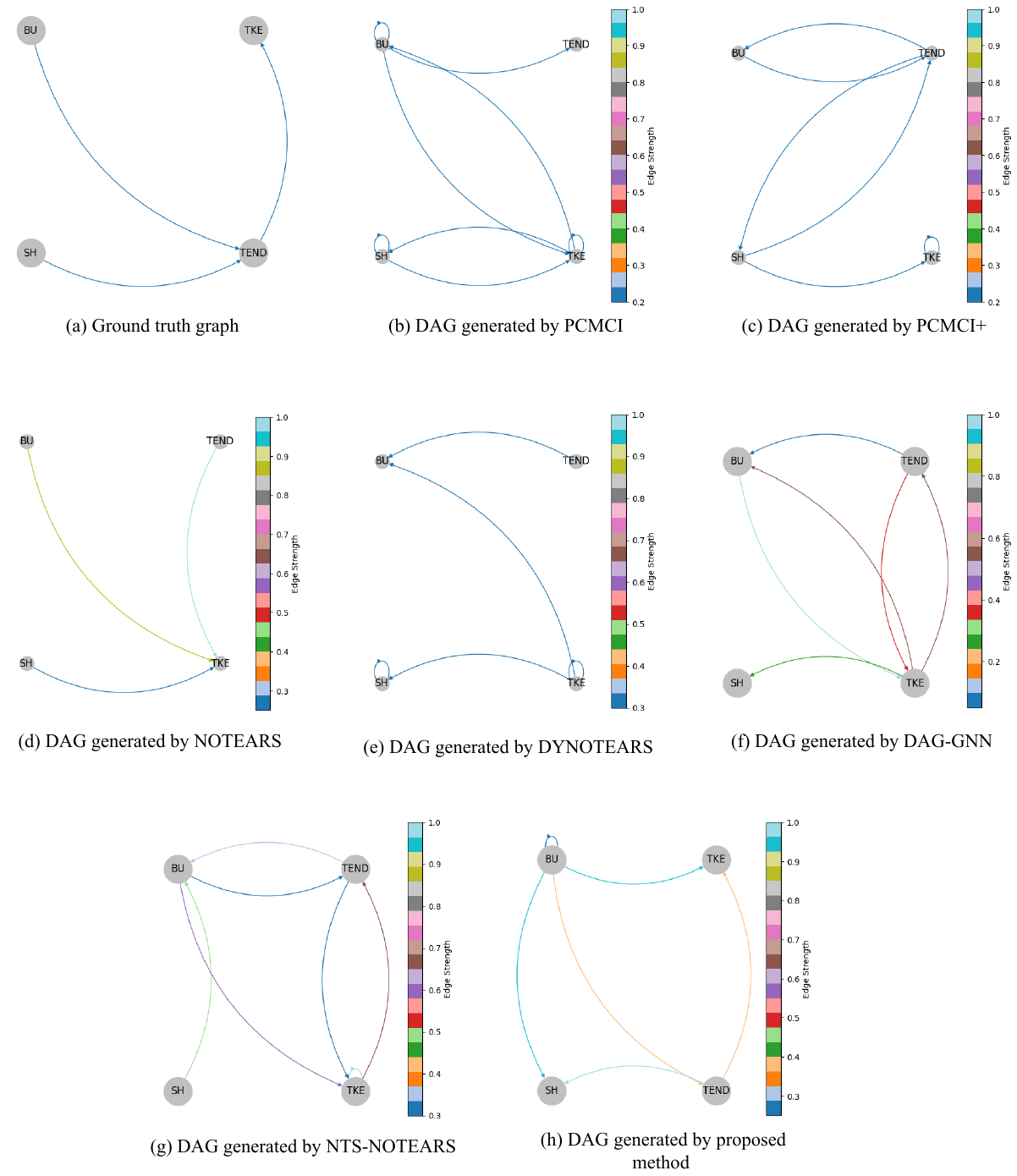}}
\caption{Visualization of summary causal graphs generated by different methods for the TKE dataset.}
\label{all-summary-plot-tke}
\end{center}
\vskip -0.2in
\end{figure*}

For the TKE dataset, the ground truth causal graph, Figure~\ref{all-summary-plot-tke}a, refers to the domain knowledge based graph generated with the aid of domain scientists. Here, the most important relationship is between $TEND$ and $TKE$ which is a unidirectional causal relation. Observing the predicted causal graphs from baseline methods, we see that the methods either miss predicting this important edge or predict a bidirectional relation between $TEND$ and $TKE$, with an exception of NOTEARS which correctly identifies this relation. Overall, our method and NOTEARS show promising results with NOTEARS having the lowest mis-prediction while our method succeeds in capturing the most important causal relation $TEND \rightarrow TKE$, highlighting its potential to make meaningful scientific contributions.

\begin{figure*}[ht]
\vskip 0.2in
\begin{center}
\centerline{\includegraphics[width=0.8\textwidth]{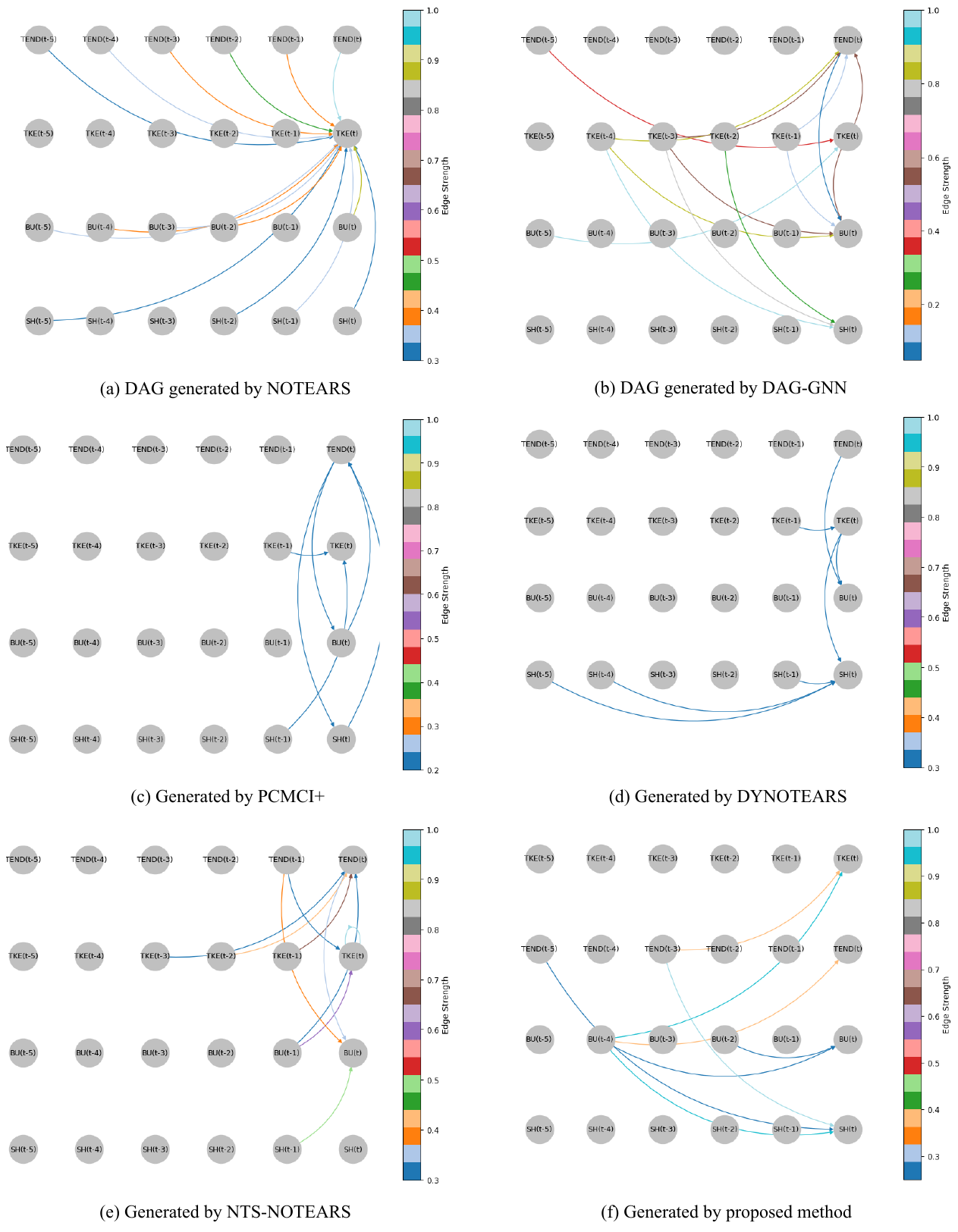}}
\caption{Visualization of full temporal causal graphs generated by different methods for the TKE dataset with time lag 5.}
\label{full-plot-tke}
\end{center}
\vskip -0.2in
\end{figure*}

\section{More Results on the Sea Ice Data}
\begin{figure*}[ht]
\vskip 0.2in
\begin{center}
\centerline{\includegraphics[width=0.7\textwidth]{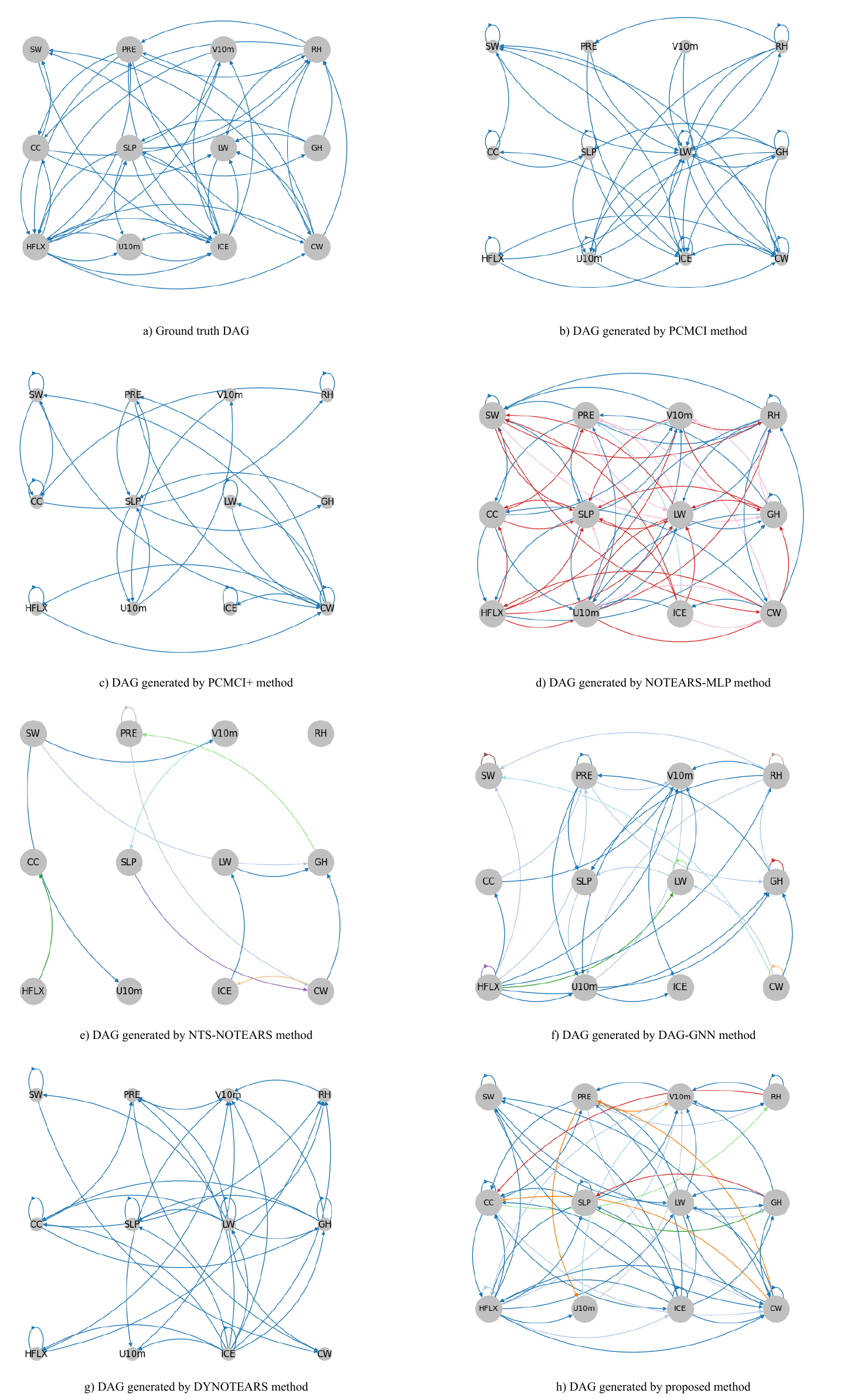}}
\caption{Visualization of summary causal graphs generated by different methods for the Arctic Sea Ice dataset with time lag 12.}
\label{summary-plot-sea-ice}
\end{center}
\vskip -0.2in
\end{figure*}

Here we present the causal graphs predicted by the proposed and baseline models on the Arctic Sea Ice dataset with a time lag of 12, referring to the annual 12-month periods respectively. We use the abbreviated terms in the causal graphs, whereas, the full form of each of the data variables is given in Table~\ref{tab:ice-data}. The true domain graph is given in Figure~\ref{summary-plot-sea-ice}a.

From the predicted graphs, we can see that NTS-NOTEARS predicted the lowest number of edges (one reason behind its low SHD), however, amongst the predicted edges, the false detection rate (FDR) is very high, for instance $ICE \leftarrow RH$. Likewise, NOTEARS-MLP predicted the highest number of edges with comparable overall performance to our proposed method. 

Since the total number of edges in the predicted and domain causal graph is very high, here, we will focus only on the edges corresponding to the scientific problem of identifying causes of sea ice melt. In this regard, we see NOTEARS-MLP, DYNOTEARS, NTS-NOTEARS and PCMCI+, fail to identify any correct causes (incoming edge) of $ICE$, whereas PCMCI, PCMCI+, DAG-GNN fail to identify any correct effect (outgoing edge) of $ICE$. Our proposed method correctly predicts 3 out of 4 outgoing edges from $ICE$, further highlighting its potential to discover complex real world causal relations.

\label{sie-all}

\end{document}